\documentclass[hidelinks,conference]{IEEEtran}

\usepackage{times}

\usepackage[numbers]{natbib}
\usepackage{multicol}
\usepackage[bookmarks=true]{hyperref}

\usepackage[utf8]{inputenc} 
\usepackage[T1]{fontenc}    
\usepackage{url}            
\usepackage{booktabs}       
\usepackage{nicefrac}       
\usepackage{microtype}      
\usepackage{hyperref}
\usepackage{amsmath,amssymb,amsfonts,amsthm}
\usepackage[export]{adjustbox}
\usepackage{caption}
\usepackage{subcaption}
\usepackage{algorithm}
\usepackage[noend]{algpseudocode}
\usepackage{xspace}
\usepackage{tabulary}
\usepackage{multirow}
\usepackage{csquotes}
\usepackage{colortbl}
\usepackage{comment}
\usepackage{soul}
\usepackage{tikz}

\newcolumntype{P}[1]{>{\centering\arraybackslash}m{#1}}

\definecolor{lightred}{rgb}{1,0.8,0.8}
\definecolor{lightgreen}{rgb}{0.8,1,0.8}

\makeatletter
\def\@IEEEsectpunct{.\ }

\def\paragraph{\@startsection{paragraph}{4}{\z@}{1.5ex plus
0.5ex minus .2ex}{-1em}{\normalsize\bf}}
\makeatother

\newcommand{\algName}[0]{\textsc{BRPO}\xspace}
\newcommand{\algFullName}[0]{Bayesian Residual Policy Optimization\xspace}

\newcommand{\algUPMLE}[0]{\textsc{UP-MLE}\xspace}

\newcommand{\algBPO}[0]{\textsc{BPO}\xspace}

\newcommand{\envMaze}[0]{\texttt{Maze4}\xspace}
\newcommand{\envMazeTen}[0]{\texttt{Maze10}\xspace}
\newcommand{\envCartpole}[0]{\texttt{Cartpole}\xspace}
\newcommand{\envWamShelf}[0]{\texttt{ArmShelf}\xspace}
\newcommand{\envCrosswalk}[0]{\texttt{CrowdNav}\xspace}

\newcommand{\envDoor}[0]{\texttt{Door4}\xspace}

\newcommand{\state}[0]{s}
\newcommand{\stateSpace}[0]{S}

\newcommand{\action}[0]{a}
\newcommand{\actionSpace}[0]{A}

\newcommand{\latent}[0]{\phi}
\newcommand{\latentSpace}[0]{\Phi}

\newcommand{\transFn}[0]{T}
\newcommand{\transFnR}[0]{T_r}

\newcommand{\policy}[0]{\pi}
\newcommand{\policyparam}[0]{r}

\newcommand{\belief}[0]{b}
\newcommand{\beliefSpace}[0]{B}

\newcommand{\rewardFn}[0]{R}
\newcommand{\reward}[0]{r}

\newcommand{\MDP}[0]{\mathcal{M}}
\newcommand{\rMDP}[0]{\mathcal{M}_r}

\newcommand{\normalization}[0]{\eta}
\newcommand{\bayesFn}[0]{\psi}\newcommand{\discount}[0]{\gamma}
\newcommand{\traj}[0]{\tau}
\newcommand{\horizon}[0]{T}

\newcommand{\expertPolicy}[0]{\pi_{e}}
\newcommand{\expert}[0]{e}
\newcommand{\residual}[0]{r}

\definecolor{SeaGreen}{HTML}{3FBC9D}
\definecolor{Orange}{HTML}{D95F02}
\definecolor{RacingGreen}{HTML}{004225}

\newcommand{\xxnote}[3]{}
\ifx\hidenotes\undefined
  \usepackage{color}
  \renewcommand{\xxnote}[3]{\color{#2}{#1: #3}}
\fi

\newcommand{\eref}[1]{(\ref{#1})}
\newcommand{\sref}[1]{Section~\ref{#1}}
\newcommand{\tabref}[1]{Table~\ref{#1}}
\newcommand{\appsref}[1]{Appendix~\ref{#1}}

\usepackage{amsmath,amsfonts,bm}

\newtheorem{theorem}{Theorem}
\newtheorem{lemma}[theorem]{Lemma}

\newenvironment{Proof}{\noindent{\emph {Proof:} }}{\qed}

\def\figref#1{Figure~\ref{#1}}
\def\Figref#1{Figure~\ref{#1}}

\def\eqref#1{equation~\ref{#1}}
\def\Eqref#1{Equation~\ref{#1}}

\def\Algref#1{Algorithm~\ref{#1}}

\def\1{\bm{1}}

\DeclareMathAlphabet{\mathsfit}{\encodingdefault}{\sfdefault}{m}{sl}
\SetMathAlphabet{\mathsfit}{bold}{\encodingdefault}{\sfdefault}{bx}{n}

\DeclareMathOperator*{\argmax}{arg\,max}

\title{\algFullName: \\ \Large Scalable Bayesian Reinforcement Learning with Clairvoyant Experts}

\author{Gilwoo Lee, Brian Hou, Sanjiban Choudhury, Siddhartha S. Srinivasa \\
Paul G. Allen School of Computer Science \& Engineering\\
University of Washington\\
\texttt{\{gilwoo,bhou,sanjibac,siddh\}@cs.uw.edu} \\
}

\begin{document}

\maketitle

\begin{abstract}
Informed and robust decision making in the face of uncertainty is critical for robots that perform physical tasks alongside people.
We formulate this as Bayesian Reinforcement Learning over latent Markov Decision Processes (MDPs).
While Bayes-optimality is theoretically the gold standard, existing algorithms do not scale well to continuous state and action spaces.
Our proposal builds on the following insight: in the absence of uncertainty, each latent MDP is easier to solve.
We first obtain an ensemble of experts, one for each latent MDP, and fuse their advice to compute a baseline policy.
Next, we train a Bayesian residual policy to improve upon the ensemble's recommendation and learn to reduce uncertainty.
Our algorithm, \algFullName (\algName), imports the scalability of policy gradient methods and task-specific expert skills.
\algName significantly improves the ensemble of experts and drastically outperforms existing adaptive RL methods.
\end{abstract}


\section{Introduction}\label{sec:intro}

\begin{figure}[t!]
\centering
\includegraphics[width=0.9\linewidth]{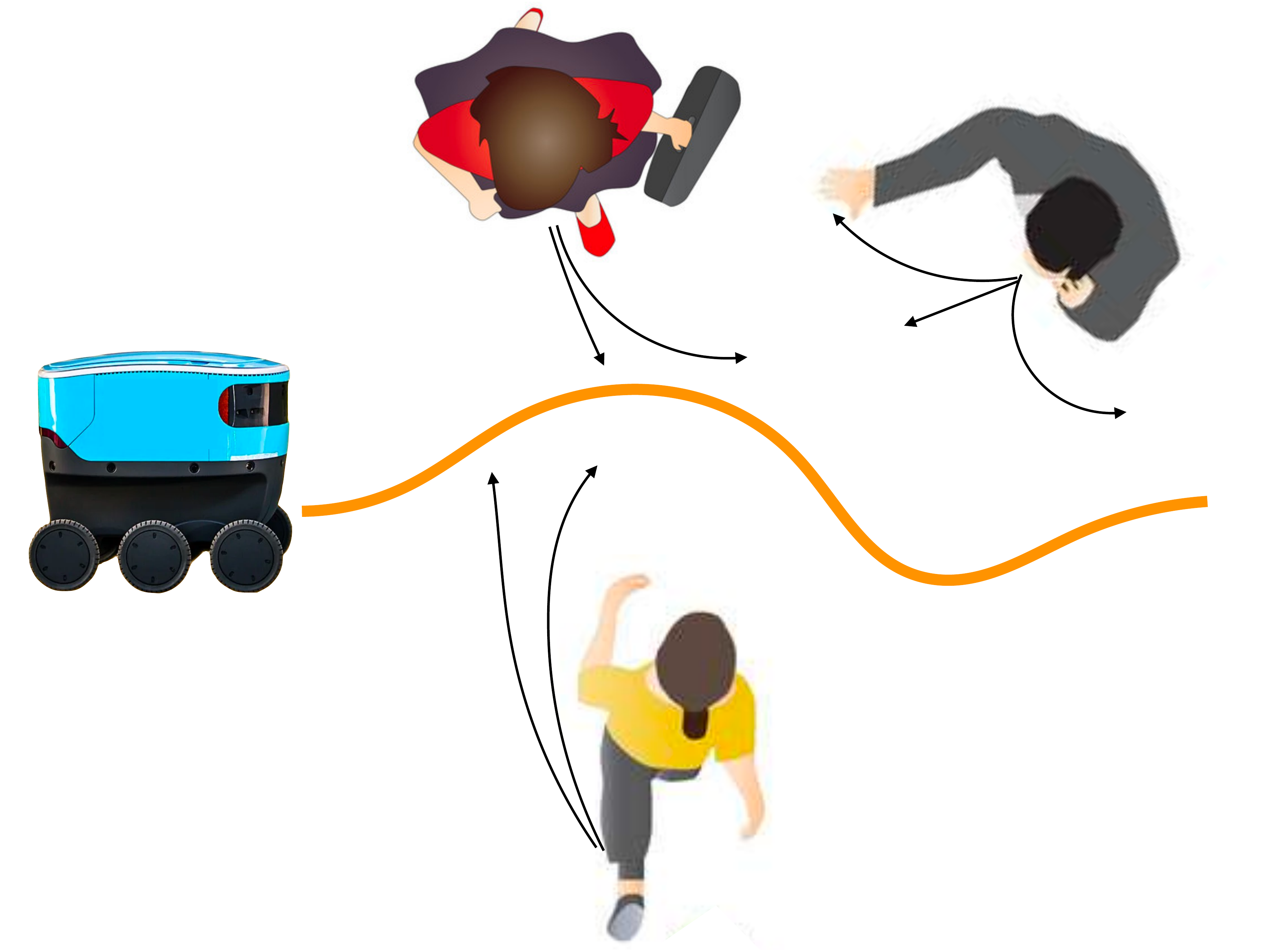}
\caption{
An autonomous vehicle approaches an area with unpredictable pedestrians, each noisily moving toward their own latent goal.
Given uncertainty on their goals, the agent must take the Bayes-optimal action to quickly drive past them without collisions.
}
\label{fig:crosswalk}
\vspace{-1em}
\end{figure}

Robots that are deployed in the real world must continue to operate in the face of model uncertainty.
For example, an autonomous vehicle must safely navigate around pedestrians navigating to latent goals (\figref{fig:crosswalk}).
A robot arm must reason about occluded objects when reaching into a cluttered shelf.
This class of problems can be framed as Bayesian reinforcement learning (BRL) where the agent maintains a belief over latent Markov Decision Processes (MDPs).
Under model uncertainty, agents do not know which latent MDP they are interacting with, preventing them from acting optimally with respect to that MDP.
At best, they can be {\it Bayes optimal}, or optimal with respect to their current uncertainty over latent MDPs.

In this work, we focus on continuous control tasks with model uncertainty.
More specifically, we aim for \emph{one-shot} test performance.
In this setting, there is no acclimation period where the agent can freely interact with the environment without consequence;
the agent will immediately accumulate rewards and incur penalties.
Thus, it must be able to address its uncertainty about the world directly, either by sensing to reduce uncertainty or selecting actions that are robust to it.

\begin{figure*}[t!]
\centering
\begin{subfigure}[b]{0.9\linewidth}
\centering
\includegraphics[width=1\linewidth]{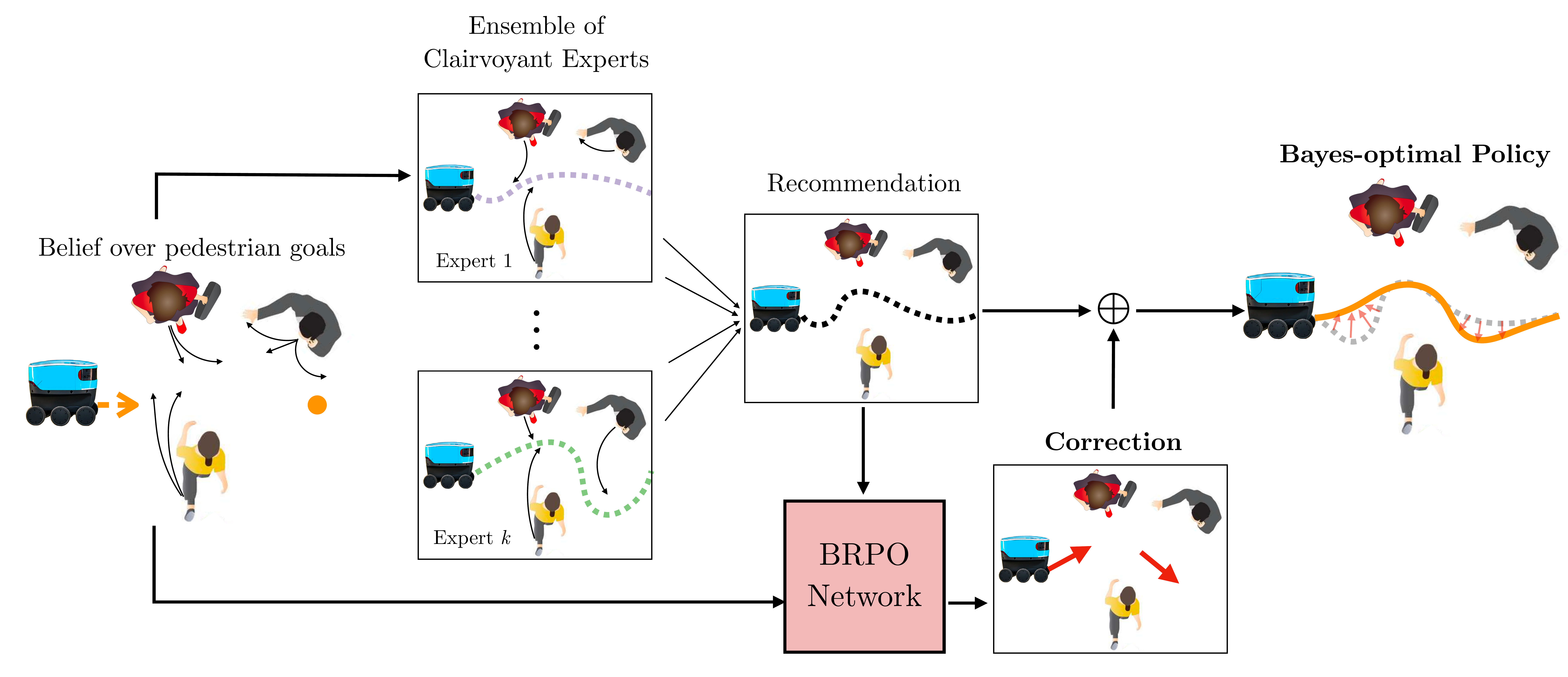}

\end{subfigure}

\caption{
  An overview of \algFullName.
  (a) Pedestrian goals are latent and tracked as a belief distribution.
  (b) Experts propose their solutions for a scenario, which are combined into a mixture of experts.
  (c) Residual policy takes in the belief and ensemble of experts' proposal and returns a correction to the proposal.
  (d) The combined \algName and ensemble policy is (locally) Bayes-optimal.
}
\label{fig:overview}
\vspace{-1em}
\end{figure*}

A Bayesian RL problem can be viewed as solving a large continuous belief MDP, which is computationally infeasible to solve directly~\citep{ghavamzadeh2015bayesian}.
These tasks, especially with continuous action spaces, are challenging for belief-space planning and robust RL algorithms.
Continuous action spaces are challenging for existing POMDP algorithms, which are either limited to discrete action spaces~\citep{kurniawati2008sarsop} or rely on online planning and samples from the continuous action space~\citep{guez2012efficient}.
Latent MDPs can be complex and may require vastly different policies to achieve high reward;
robust RL methods~\citep{rajeswaran2016epopt,tobin2017domain} are often unable to produce that multi-modality.

We build upon a simple yet recurring observation~\citep{choudhury2018data,kahn2017plato,osband2013psrl}:
while solving the belief MDP is hard, solving individual latent MDPs is much more tractable.
If the path for each pedestrian is known, the autonomous vehicle can invoke a motion planner that avoids collision.
We can think of these solutions as \emph{clairvoyant experts}, i.e., experts that think they know the latent MDP and offer advice accordingly.
Combining advice from the clairvoyant experts can be effective, but such an ensemble policy can struggle with model uncertainty.
Since experts are individually confident about which MDP the agent faces, the ensemble will never prioritize uncertainty reduction or robust actions, which may be valuable in the long run.

Our algorithm, \algFullName (\algName), computes a {\it residual} policy to augment a given ensemble of clairvoyant experts (\figref{fig:overview}).
This is computed via policy optimization in a residual belief MDP,
induced by the ensemble policy's actions on the original belief MDP.
Because the ensemble is near-optimal when the entropy is low, \algName can focus on learning how to safely collapse uncertainty in regions of higher entropy.
It can also start with much higher performance than when starting from scratch, which we prove in \sref{sec:algorithm} and empirically validate in \sref{sec:results}.

Our key contribution is the following:
\begin{itemize}
\item We propose \algName, a scalable Bayesian RL algorithm for one-shot performance on problems with model uncertainty.
\item We prove that it monotonically improves upon the expert ensemble, converging to a locally Bayes-optimal policy.
\item We experimentally demonstrate that \algName outperforms both the ensemble and existing adaptive RL algorithms.
\end{itemize}


\section{Related Work}

\paragraph{POMDP methods}
Bayesian reinforcement learning formalizes RL where one has a prior distribution over possible MDPs~\citep{ghavamzadeh2015bayesian,shani2013survey}. However, the Bayes-optimal policy, which is the best one can do under uncertainty, is intractable to solve for and approximation is necessary~\citep{hsu2008makes}.
One way is to approximate the value function, as done in SARSOP~\citep{kurniawati2008sarsop} and PBVI~\citep{pineau2003point}; however, they cannot deal with continous state actions. Another strategy is to resort to sampling, such as BAMCP~\citep{guez2012efficient}, POMCP~\citep{silver2010monte}, POMCPOW~\citep{sunberg2018pomcpow}. However, these approaches require a significant amount of online computation.

Online approaches forgo acting Bayes-optimally right from the onset, and instead aim to eventually act optimally. The question then becomes: how do we efficiently gain information about the test time MDP to act optimally? BEB~\citep{kolter2009near} and POMDP-lite~\citep{chen2016pomdp} introduce an auxiliary reward term to encourage exploration and prove Probably-Approximately-Correct~(PAC) optimality. This has inspired work on more general, non-Bayesian curiosity based heuristics for reward gathering~\citep{achiam2017surprise,Burda2018a,houthooft2016vime,pathak2017curiosity}. Online exploration is also well studied in the bandit literature, and techniques such as posterior sampling~\citep{JMLR:v20:18-339} bound the learner's regret. UP-OSI~\citep{yu2017uposi} predicts the most likely MDP and maps that to an action. \citet{gimelfarb2018reinforcement} learns a gating over multiple expert value functions. However, online methods can over-explore to unsafe regimes.


Another alternative is to treat belief MDP problems as a large state space that must be compressed. \citet{peng2018sim} use Long Short-Term Memory~(LSTM)~\citep{hochreiter1997long} to encode a history of observations to generate an action. Methods like BPO~\citep{lee2018bayesian} explicitly utilize the belief distribution and compress it to learn a policy. The key difference between \algName and BPO is that \algName uses an expert, enabling it to scale to handle complex latent tasks that may require multimodal policies.


\paragraph{Meta-reinforcement Learning}
Meta-reinforcement learning (MRL) approaches train sample-efficient learners by exploiting structure common to a distribution of MDPs. For example, MAML~\citep{finn2017model} trains gradient-based learners while RL2~\citep{duan2016rl} trains memory-based learners.
While meta-supervised learning has well established Bayesian roots~\citep{baxter1998theoretical,baxter2000model}, it wasn't until recently that meta-reinforcement learning was strongly tied to Bayesian Reinforcement Learning (BRL)~\citep{ortega2019meta,rabinowitz2019meta}.
Nevertheless, even non-Bayesian MRL approaches address problems pertinent to BRL.
MAESN~\citep{gupta2018meta} learns structured noise for exploration.
E-MAML~\citep{stadie2018meta} adds an explicit exploration bonus to the MAML objective.
GMPS~\citep{mendonca2019guided} exploit availability of MDP experts to partially reduce BRL to IL.
Our work is more closely related to Bayesian MRL approaches.
MAML-HB~\citep{grant2018recasting} casts MAML as hierarchical Bayes and improves posterior estimates.
BMAML~\citep{yoon2018bayesian} uses non-parametric variational inference to improve posterior estimates.
PLATIPUS~\citep{finn2018promaml} learns a parameter distribution instead of a fixed parameter. PEARL~\citep{rakelly2019efficient} learns a data-driven Bayes filter across tasks.
In contrast to these approaches, we use experts at test time, learning only to optimally correct them.

\paragraph{Residual Learning}
Residual learning has its foundations in boosting~\citep{freund1999short} where a combination of weak learners, each learning on the failures of previous, make a strong learner. It also allows for injecting priors in RL, by boosting off of hand-designed policies or models. Prior work has leveraged known but approximate models by learning the residual between the approximate dynamics and the discovered dynamics~\citep{ostafew2014learning,ostafew2015conservative,berkenkamp2015safe}. There has also been work on learning residual policies over hand-defined ones for solving long horizon~\citep{silver2018residual} and complex control tasks~\citep{johannink2019residual}. Similarly, our approach starts with a useful initialization (via experts) and learns to improve via Bayesian reinforcement learning.

\section{Preliminaries: Bayesian Reinforcement Learning}
\label{sec:prelim}


We are interested in one-shot performance of an RL agent under model uncertainty, in which the latent model gets reset at the beginning of each episode.
As discussed in \sref{sec:intro}, this problem can be formulated as model-based Bayesian Reinforcement Learning~(BRL).
Formally, the problem is defined by a tuple $\langle \stateSpace,\latentSpace, \actionSpace, \transFn, \rewardFn, P_0, \discount \rangle$, where $\stateSpace$ is the observable state space of the underlying MDPs, $\latentSpace$ is the latent space, and $\actionSpace$ is the action space.
$\transFn$ and $\rewardFn$ are the transition and reward functions parameterized by $\latent$.
The initial distribution over $(\state, \latent)$ is given by $P_0: \stateSpace \times \latentSpace \rightarrow \mathbb{R}^+$, and $\discount$ is the discount.

Since the latent variable is not observable, Bayesian RL considers the long-term expected reward with respect to the uncertainty over $\latent$ rather than the true (unknown) value of $\latent$.
Uncertainty is represented as a \emph{belief distribution} $\belief \in \beliefSpace$ over latent variables $\latent$.
The Bayes-optimal action value function is given by the Bellman equation:
\begin{equation}\label{eq:rl}
\begin{aligned}
Q(\state, \belief, \action)
&= \rewardFn(\state, \belief, \action) \\
&+
\discount \sum_{\state', \belief'} P(\state', \belief' | \state, \belief, \action) \max_{\action'} Q(\state', \belief', \action')
\end{aligned}
\end{equation}
where $R(s, b, a)
  = \sum_{\latent \in \latentSpace} \belief(\latent) \rewardFn(\state, \latent, \action)$ and
$P(s'| s, b, a)
= \sum_{\latent\in\latentSpace} \belief(\latent) P(\state' | \state, \latent, \action)$.
The posterior update $P(\belief' | \state, \belief, \action)$ is computed recursively, starting from initial belief $\belief_0$.
$\belief'(\latent' | \state, \belief, \action, \state')
  = \normalization \sum_{\latent \in \latentSpace} \belief(\latent)
  \transFn(\state, \latent, \action, \state', \latent')$ 
where $\normalization$ is the normalizing constant, and the transition function is defined as $\transFn(\state, \latent, \action, \state', \latent')
= P(\state' | \state, \latent, \action) P(\latent' | \state, \latent, \action, \state')$.

While some terminology is shared with online RL algorithms (e.g. Posterior Sampling Reinforcement Learning~\citep{osband2013psrl}), that setting assumes latent variables are fixed for multiple episodes.
We refer the reader to \appsref{ssec:psrl} for further discussion.


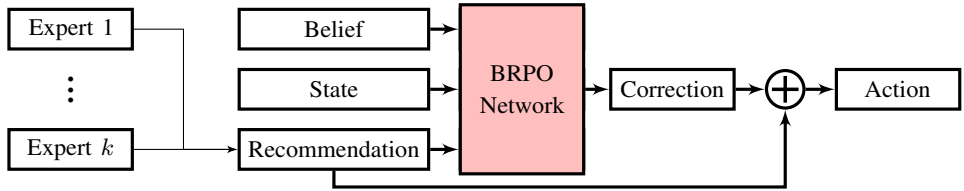
\begin{figure*}[t!]
\centering
\usetikzlibrary{shapes,arrows}
\tikzset{>=latex}
\tikzset{every node/.style={align=center}}

\tikzstyle{block} = [draw, very thick, rectangle, text width=4.0em, text centered, minimum height=1.5em, fill=white]
\tikzstyle{cloud} = [draw, very thick, circle, text width=1em, text centered, inner sep=0pt, minimum height=1em]
\tikzstyle{line}  = [draw, very thick, -latex']


\begin{tikzpicture}[auto]

  \node [block] at (-6,  0.8) (expert-1) {{\small Expert 1}};
  \node []      at (-6,  0.1) (dots)     {\LARGE \vdots};
  \node [block] at (-6, -0.8) (expert-k) {{\small Expert $k$}};

  \node [block, text width=6.5em] at (-2.5, 0.8) (belief) {\small Belief};
  \node [block, text width=6.5em] at (-2.5, 0) (state) {\small State};
  \node [block, text width=6.5em] at (-2.5, -0.8) (expert) {{\small Recommendation}};

  \node [block] at (0,  0.8) (belief-target) {};
  \node [block] at (0,  0.0) (state-target)  {};
  \node [block] at (0, -0.8) (expert-target) {};

  \node [block, fill=pink, minimum height=6.4em] at (0, 0) (policy-network) {{\small BRPO\\Network}};
  \node [block] at (2, 0) (residual-action) {{\small Correction}};
  \node [cloud] at (3.5, 0) (sum) {\huge +};
  \node [block] at (5, 0) (action) {{\small Action}};

  \path [line] (policy-network) -- (residual-action);
  \path [line] (residual-action) -- (sum);
  \path [line] (expert) -- ++(0, -0.5) -| (sum);
  \path [line] (sum) -- (action);
  \path [line] (belief) -- (belief-target);
  \path [line] (state)  -- (state-target);
  \path [line] (expert) -- (expert-target);

  \path [draw, thin, -latex'] (expert-1) -- ++(1.5, 0) |- (expert);
  \path [draw, thin, -latex'] (expert-k) -- (expert);
\end{tikzpicture}
\caption{
  Bayesian residual policy network architecture.
}
\label{fig:network}
\end{figure*}

\begin{algorithm}[!tb]
\caption{\algFullName}
\label{algo:rbpo}
\begin{algorithmic}[1]
\Require Bayes filter $\bayesFn$, belief $\belief_0$, prior $P_0$, residual policy $\policy_{\policyparam_0}$, expert $\expertPolicy$, horizon $\horizon$, $n_\text{itr}, n_\text{sample}$
\Statex
\For{$i = 1, 2, \cdots, n_\text{itr}$}
  \For{$n = 1, 2 , \cdots, n_\text{sample}$}
    \State Sample latent MDP $\MDP$: $(\state_0, \latent_0) \sim P_0$
    \State $\traj_n \leftarrow \texttt{Simulate}(\policy_{\policyparam_{i-1}}, \expertPolicy, \belief_0, \bayesFn, \MDP, \horizon)$
  \vspace{-1mm}
  \EndFor
  \State $\policy_{\policyparam_i} \leftarrow \texttt{BatchPolicyOpt}(\policy_{\policyparam_{i-1}}, \{\traj_n\}_{n=1}^{n_\text{sample}})$
\vspace{-1mm}
\EndFor
\State \Return $\policy_{\policyparam_{best}}$
\Statex
\Procedure{Simulate}{$\policy_{\policyparam}, \expertPolicy, \belief_0, \bayesFn, \MDP, \horizon$}
\For {$t = 1, \cdots, \horizon$}
\State $\action_{\expert_t} \sim  \expertPolicy(\state_t, \belief_t)$  // Expert recommendation \label{alg:expert-line}
\State $\action_{\residual_t} \sim \policy_{\policyparam} (\state_{t}, \belief_{t}, \action_{\expert_t})$ // Residual policy \label{alg:residual-line}
\State $\action_t \leftarrow \action_{\residual_t}+ \action_{\expert_t}$
\State Execute $\action_t$ on $\MDP$, receive $\reward_{t+1}$, observe $\state_{t+1}$ \label{alg:execution}
\State $\belief_{t+1} \leftarrow \bayesFn(\state_{t}, \belief_{t}, \action_{t}, \state_{t+1})$ // Belief update \label{alg:belief-update}
\vspace{-1mm}
\EndFor
\State  $\traj \leftarrow (\state_0, \belief_0, \action_{\residual_0}, \reward_1, \state_1, \belief_1, \cdots, \state_\horizon, \belief_\horizon)$ \label{alg:traj-line}
\State \Return $\traj$
\EndProcedure
\end{algorithmic}
\end{algorithm}

\section{\algFullName (\algName)}
\label{sec:algorithm}

\algFullName relies on an ensemble of clairvoyant experts where each expert solves a latent MDP.
This is a flexible design parameter with three guidelines.
First, the ensemble must be fixed before training begins.
This fixes the residual belief MDP, which is necessary for theoretical guarantees (\sref{ssec:brpo-guarantee}).
Next, the ensemble should return its recommendation quickly since it will be queried online at test time.
Practically, we have observed that this factor is often more important than the strength of the initial ensemble:
even weaker ensembles can provide enough of a head start for residual learning to succeed.
Finally, when the belief has collapsed to a single latent MDP, the resulting recommendation must follow the corresponding expert.
In general, the ensemble should become more reliable as entropy decreases.


\algName performs batch policy optimization in the residual belief MDP, producing actions that continuously correct the ensemble recommendations.
Intuitively, \algName enjoys improved data-efficiency because the correction can be small when the ensemble is effective (e.g., when uncertainty is low or when the experts are in agreement).
When uncertainty is high, the agent learns to override the ensemble, reducing uncertainty and taking actions robust to model uncertainty.

\subsection{Ensemble of Clairvoyant Experts}\label{ssec:ensemble}

For simplicity of exposition, assume the Bayesian RL problem consists of $k$ underlying latent MDPs, $\latent_1, ..., \latent_k$.
The ensemble policy maps the state and belief to a distribution over actions $\policy_e: \stateSpace \times  \beliefSpace \rightarrow P(\actionSpace)$.
It combines clairvoyant experts $\policy_1, \cdots, \policy_k$, one for each latent variable $\latent_i$.
Each expert can be computed via single-MDP RL (or optimal control, if transition and reward functions are known by the experts).

There are various strategies to produce an ensemble from a set of experts.
The ensemble $\policy_e$ could be the maximum a posteriori (MAP) expert: $\pi_e = \argmax_{b(\phi)} \pi_\phi$.
This particular ensemble allows \algName to solve tasks with infinitely many latent MDPs, as long as the MAP expert can be queried online.
It can also be a weighted sum of expert actions, which turns out to be the MAP action for Gaussian policies~(\appsref{ssec:appendix-map-expert}).


While these belief-aware ensembles are easy to attain, they are not Bayes-optimal.
In particular, since each clairvoyant expert assumes a perfect model, the ensemble does not take uncertainty reducing actions nor is it robust to model uncertainty.
Instead of constructing an ensemble of experts, one could approximately solve the BRL problem with a POMDP solver and perform residual learning on this policy.
While the initial policy would be much improved, state-of-the-art online POMDP solvers are expensive and would be slow at test time.

\subsection{Bayesian Residual Policy Learning}\label{ssec:brpo-algo}

In each training iteration, \algName collects trajectories by simulating the current policy on several MDPs sampled from the prior distribution (\Algref{algo:rbpo}).
At every timestep of the simulation, the ensemble is queried for an action recommendation~(Line~\ref{alg:expert-line}), which is summed with the correction from the residual policy network (\figref{fig:network}) and executed~(Line \ref{alg:residual-line}-\ref{alg:execution}).
The Bayes filter updates the posterior after observing the resulting state~(Line~\ref{alg:belief-update}).
The collected trajectories are the input to a policy optimization algorithm, which updates the residual policy network.
Note that only {\it residual actions} are collected in the trajectories~(Line~\ref{alg:traj-line}).

The \algName agent effectively experiences a different MDP:
in this new MDP, actions are always shifted by the ensemble recommendation.
We formalize this correspondence between the residual and original belief MDPs in the next section, enabling \algName to inherit the monotonic improvement guarantees from existing policy optimization algorithms.

\subsection{BRPO Inherits Motononic Improvement Guarantees}\label{ssec:brpo-guarantee}

In this section, we prove that \algName guarantees monotonic improvement on the expected return of the mixture between the ensemble policy $\policy_e$ and the initial residual policy $\policy_{r_0}$.
If the initial residual policy's actions are small, this is close to the expected return of the ensemble.
The following arguments apply to all MDPs, not just belief MDPs.
For clarity of exposition, we have omitted the belief from the state.

First, we observe that $\policy_r$ operates on its own residual MDP.
Because the ensemble policy $\policy_e$ is fixed, this residual MDP is fixed as well.\footnotemark
\footnotetext{
  If the ensemble changes during training, the residual cannot be cast as a single MDP.
  Proofs for policy optimization algorithms no longer hold.
}
Finally, we show that the monotonic guarantee on the residual MDP can be transferred to the original MDP.



Let $\MDP = \langle \stateSpace, \actionSpace, \transFn, \rewardFn, P_0 \rangle$ be the original MDP.
For simplicity, assume that $\rewardFn$ depends only on states.\footnotemark
\footnotetext{If $\rewardFn$ is dependent on actions, we can define $\rewardFn_r$ analogous to \eref{eq:transition-shift}.}
Every $\policy_e$ for $\MDP$ induces a residual MDP $\rMDP = \langle \stateSpace, \actionSpace_r, \transFn_r, \rewardFn, P_0 \rangle$ that is equivalent to $\MDP$ except for the action space and transition function.%
\footnote{
  $\actionSpace_r = \actionSpace$ as long as the expert action space contains the null action.
}
For every residual action $\action_{r}$, $\transFnR$ marginalizes over all expert recommendations $\action_e \sim \policy_{e}(\state)$.
\begin{align}\label{eq:transition-shift}
\transFnR &(\state' | s, a_{r})
= \sum_{\action_e} \transFn (\state' | \state, a_e + a_r) \policy_e(a_e|\state)
\end{align}
Let $\policy_r(\action_r | \state, \action_e)$ be a residual policy.
The final policy $\policy$ executed on $\MDP$ is a mixture of $\policy_r$ and $\policy_e$, since actions are sampled from both policies and summed.
\begin{equation}\label{eq:combined-policy}
\policy(\action|\state) = \sum_{\action_r} \policy_e(\action - \action_r|s)\policy_r(\action_r|\state, \action - \action_r)
\end{equation}

\begin{lemma}\label{lem:ppo-on-res}
\algName monotonically improves the expected return of $\policy_r$, i.e.,
\begin{equation*}
J(\policy_{r_{i+1}}) \geq J(\policy_{r_{i}})
\end{equation*}
where $J(\policy) = \mathbb{E}_{\tau\sim \policy}[R(\tau)]$.
\end{lemma}
\begin{Proof}
\algName uses PPO for optimization~\cite{schulman2017proximal}.
PPO's clipped surrogate objective approximates the following objective,
\begin{equation}
\max_\theta \mathbb{\hat{E}}\left[\frac{\policy_\theta(\action_t|\state_t)}{\policy_{\theta_{\text{old}}}(\action_t|\state_t)}\hat{A}_t - \beta \cdot \mathrm{KL}(\policy_{\theta_{\text{old}}} (\cdot|\state_t), \policy_\theta(\cdot|\state_t)) \right],
\end{equation}
where $\policy_\theta$ is a policy parameterized by $\theta$ and $\policy_{\theta_{\text{old}}}$ is the policy in the previous iteration, which correspond to the current and previous residual policies $\policy_{r_i}, \policy_{r_{i-1}}$ in \Algref{algo:rbpo}.
$\hat{A}$ is the generalized advantage estimate~(GAE) and KL is the Kullback–Leibler divergence between the two policy distributions.
PPO proves monotonic improvement for the policy's expected return by bounding the divergence from the previous policy in each update.
This guarantee only holds if both policies are applied to the same residual MDP, i.e. the ensemble is fixed.
\end{Proof}

Next, we show that the performance of $\policy_r$ on the residual MDP $\rMDP$ is equivalent to the \algName agent's actual performance on the original MDP $\MDP$.

\begin{theorem}\label{theorem:brpo}
A residual policy $\policy_r$ executed on $\rMDP$ has the same expected return as the mixture policy $\policy$ executed on $\MDP$.
\begin{equation*}
\mathbb{E}_{\traj\sim(\policy, \MDP)}[\rewardFn(\traj)] = \mathbb{E}_{\traj_r\sim(\policy_r, \rMDP)}[\rewardFn(\traj_r)]
\end{equation*}
$\traj\sim(\policy, \MDP)$ indicates that $\traj$ is a trajectory with actions sampled from $\policy$ and executed on $\MDP$.
\end{theorem}
We first show that the probability of observing a sequence of states is equal in both MDPs, which immediately leads to this theorem.

Let $\xi = (s_0, s_1, ..., s_{T-1})$ be a sequence of states.
Let $\alpha=\{\tau\}$ be the set of all length $T$ trajectories (state-action sequences) in $\MDP$ with $\xi$ as the states, and $\beta=\{\tau_r\}$ be analogously defined for a  set of trajectories in $\rMDP$.
Note that for each state-sequence $\xi$ may have multiple state-action trajectories $\{\tau\}$ containing $\xi$.
\begin{lemma}\label{lem:seq-prob}
The probability of $\xi$ is equal when executing $\pi$ on $\MDP$ and $\pi_r$ on $\rMDP$, i.e.,
\begin{equation*}
\pi(\xi) = \sum_{\tau\in \alpha} \pi(\tau) = \sum_{\tau_r\in\beta} \pi_r(\tau_r) = \pi_r(\xi)
\end{equation*}
\end{lemma}
\begin{Proof}
(i) Base case, $T=0$. It holds trivially since $\MDP$ and $\rMDP$ share the same initial state distribution $P_0$.\\
(ii) Assume it holds for $T=t$. Pick any $\xi$ and let its last element be $s$.
Consider an $s'$-extended sequence $\xi' = (\xi, s')$.
Conditioned on $\xi$, the probability of $\xi'$ is equal in $(\pi, \MDP)$ and $(\pi_r, \rMDP)$, which we can see by marginalizing over all state-action sequences:
\begin{align}
\sum_{\tau'_{r}} &\pi_r(\tau'_r|\xi) = \sum_{a_r}  \pi_r(a_r|s) T_r(s'|s,a_r) \label{eeq:1} \\
&=  \sum_{a_r} \pi_r(a_r|s) \sum_{a} T(s'|s, a) \pi_e(a-a_r|s) \label{eeq:2} \\
&=\sum_{a} \sum_{a_r} \pi_r(a_r|s)\pi_e(a-a_r|s)T(s'|s, a) \label{eeq:3}\\
&=\sum_{a} \pi(a|s) T(s'|s, a) \label{eeq:4} \\
&=\sum_{\tau'} \pi(\tau'|\xi)
\end{align}
The transition from \eref{eeq:1} to \eref{eeq:2} comes from \eref{eq:transition-shift} and \eref{eeq:3} to \eref{eeq:4} comes from \eref{eq:combined-policy}.
It follows that,
\begin{equation*}
\pi(\xi') = \pi(\xi) \sum_{\tau'} \pi (\tau'|\xi) = \pi_r(\xi) \sum_{\tau'_r}\pi_r(\tau_r'|\xi) = \pi_r(\xi'),
\end{equation*}
which proves the lemma.
\end{Proof}

Since reward depends only on the states,
$R(\tau) = R(\tau_r)=R(\xi)$ for all $\tau \in \alpha, \tau_r \in \beta$.
Hence, Lemma \ref{lem:seq-prob} immediately implies Theorem \ref{theorem:brpo}.
\begin{align*}
\mathbb{E}&_{\tau} [R(\tau)]=\sum_\tau R(\tau)\pi(\tau) = \sum_\xi R(\xi)\pi(\xi)\\
&= \sum_\xi R(\xi) \pi_r(\xi) = \sum_{\tau_r} R(\tau_r) \pi_r(\tau_r) = \mathbb{E}_{\tau_r} [R(\tau_r)]
\end{align*}

Finally, we prove our main theorem that the monotonic improvement guarantee on $\rMDP$ transfers to $\MDP$.

\begin{theorem}
  \algName monotonically improves upon the mixture between ensemble policy $\policy_e$ and initial residual policy $\policy_{r_0}$, eventually converging to a locally optimal policy.
\end{theorem}
\begin{Proof}
From Lemma~\ref{lem:ppo-on-res}, we have that $\policy_r$ monotonically improves on the residual MDP $\rMDP$.
From Theorem~\ref{theorem:brpo}, monotonic improvement of $\policy_r$ on $\rMDP$ implies monotonic improvement of the mixture policy $\policy$ on the actual MDP $\MDP$.
If the initial residual policy's actions are small, the expected return of the mixture policy $\policy$ on $\MDP$ is close to that of the ensemble $\policy_e$.
\end{Proof}

In summary, \algName tackles the one-shot setting for problems with model uncertainty by
building on an ensemble of clairvoyant experts (queried online)
and optimizing a policy on the residual MDP induced by the ensemble (trained offline, queried online).
Even suboptimal ensembles often provide a strong baseline, resulting in data-efficient learning and high returns.
We empirically evaluate this hypothesis in \sref{sec:results}.


\begin{figure}
    \centering
    \begin{subfigure}{0.45\linewidth}
      \centering
      \includegraphics[width=1\linewidth]{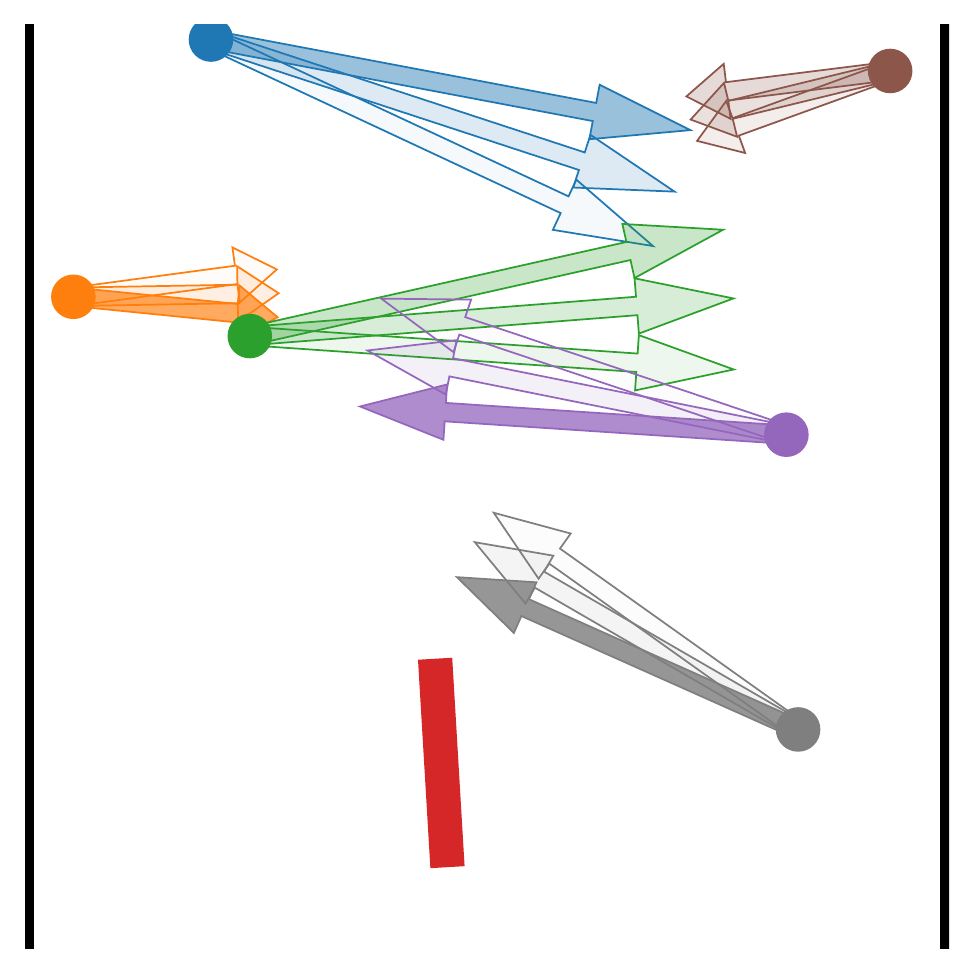}
      \caption{\envCrosswalk}
      \label{fig:crosswalk-setup}
    \end{subfigure}
    ~
    \begin{subfigure}{0.45\linewidth}
      \centering
      \includegraphics[width=1\linewidth]{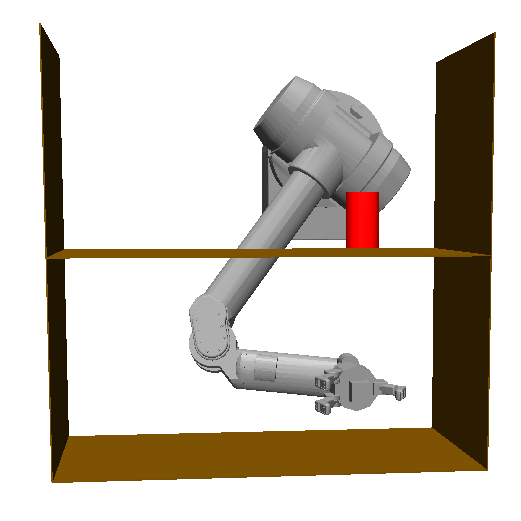}
      \caption{\envWamShelf}
      \label{fig:wam-setup}
    \end{subfigure}
    \vspace{0.5em}
    \caption{
      Setup for \envCrosswalk and \envWamShelf.
      In \envCrosswalk, the goal for the agent (red) is to go upward without colliding with pedestrians (all other colors).
      In \envWamShelf, the goal is to reach for the can under noisy sensing. See \sref{ssec:envs}.
    }
    \label{fig:cartpole-wam-setup}
\end{figure}

\section{Experimental Results}
\label{sec:results}

We choose problems that highlight common challenges for robots with model uncertainty:
\begin{itemize}
\item Costly sensing is required to infer the latent MDP.
\item Uncertainty reduction and robustness are critical.
\item Solutions for each latent MDP are significantly different.
\end{itemize}
In all domains that we consider, \algName improves on the ensemble's recommendation and significantly outperforms adaptive-RL baselines that do not leverage experts.
Qualitative evaluation shows that robust Bayes-optimal behavior naturally emerges from training.

\begin{figure*}[t!]
  \begin{subfigure}{1\linewidth}
    \centering
    \includegraphics[width=1\linewidth]{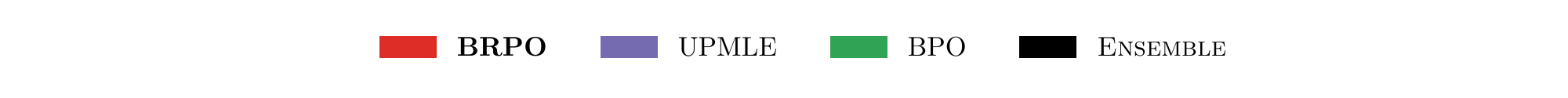}
  \end{subfigure}
  \\
    \begin{subfigure}{0.15\linewidth}
    \centering
    \includegraphics[width=1\linewidth]{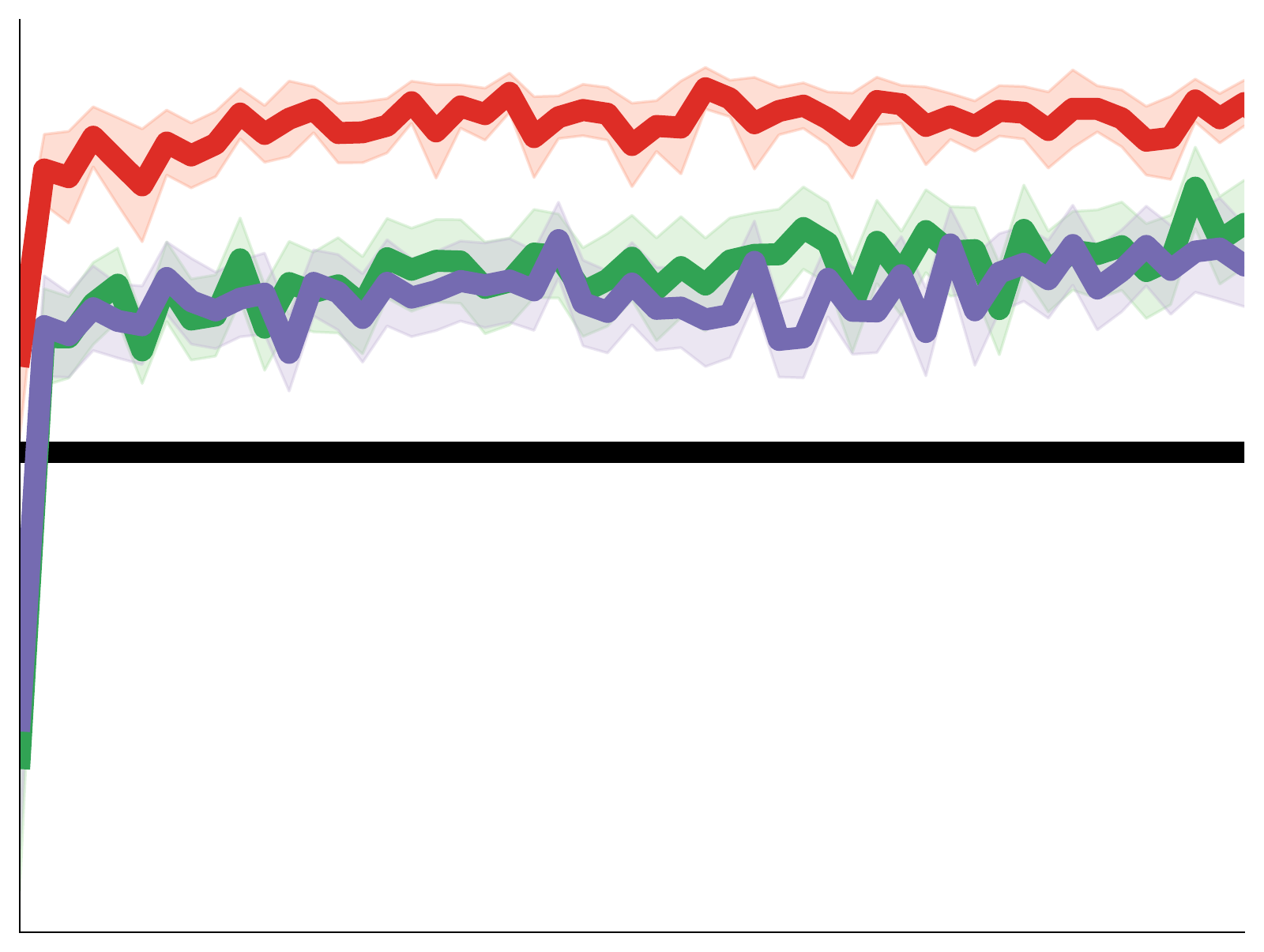}
    \caption{\envCrosswalk}
  \end{subfigure}
  ~
  \begin{subfigure}{0.15\linewidth}
    \centering
    \includegraphics[width=1\linewidth]{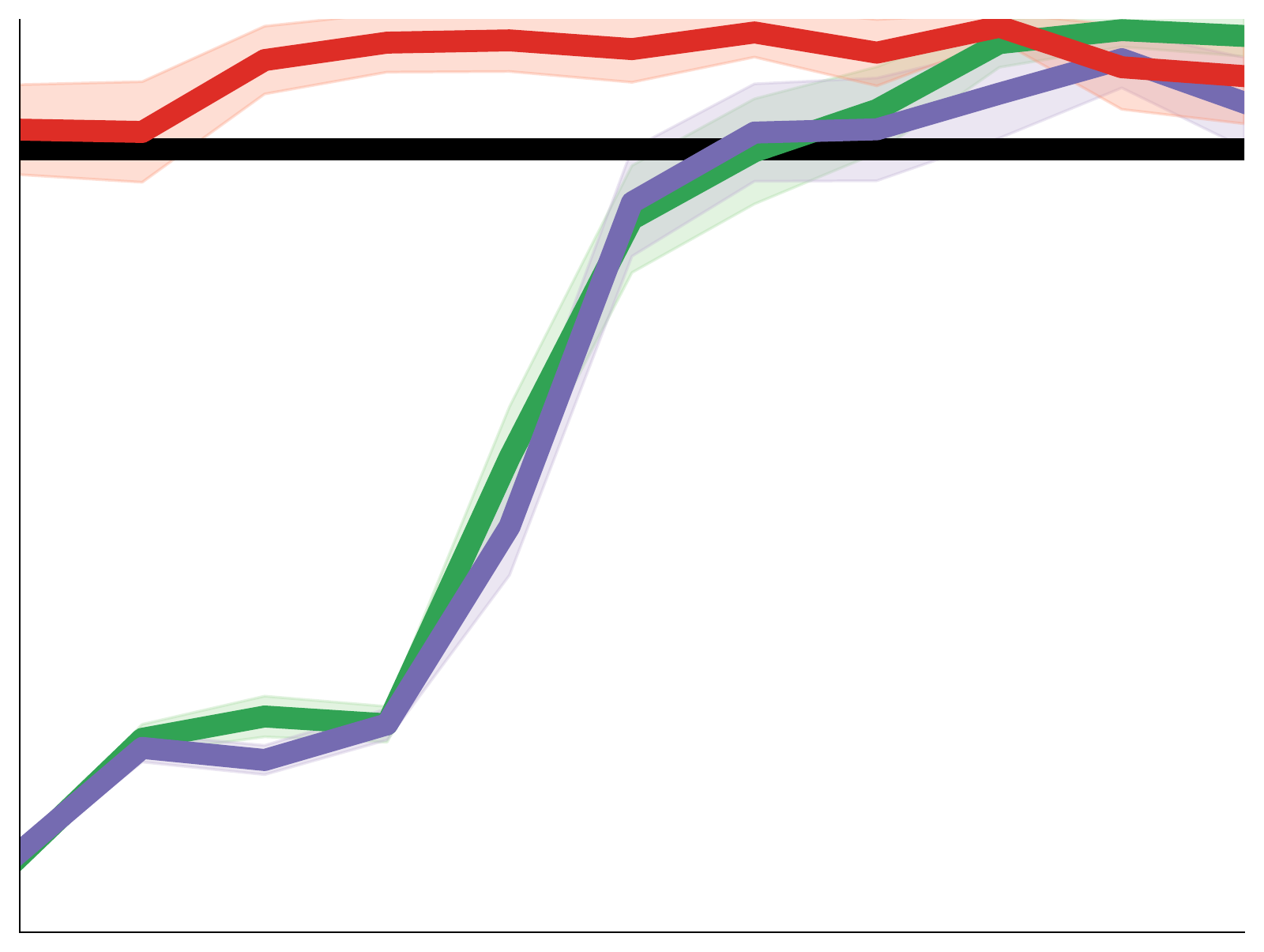}
    \caption{\envCartpole}
  \end{subfigure}
  ~
  \begin{subfigure}{0.15\linewidth}
    \centering
    \includegraphics[width=1\linewidth]{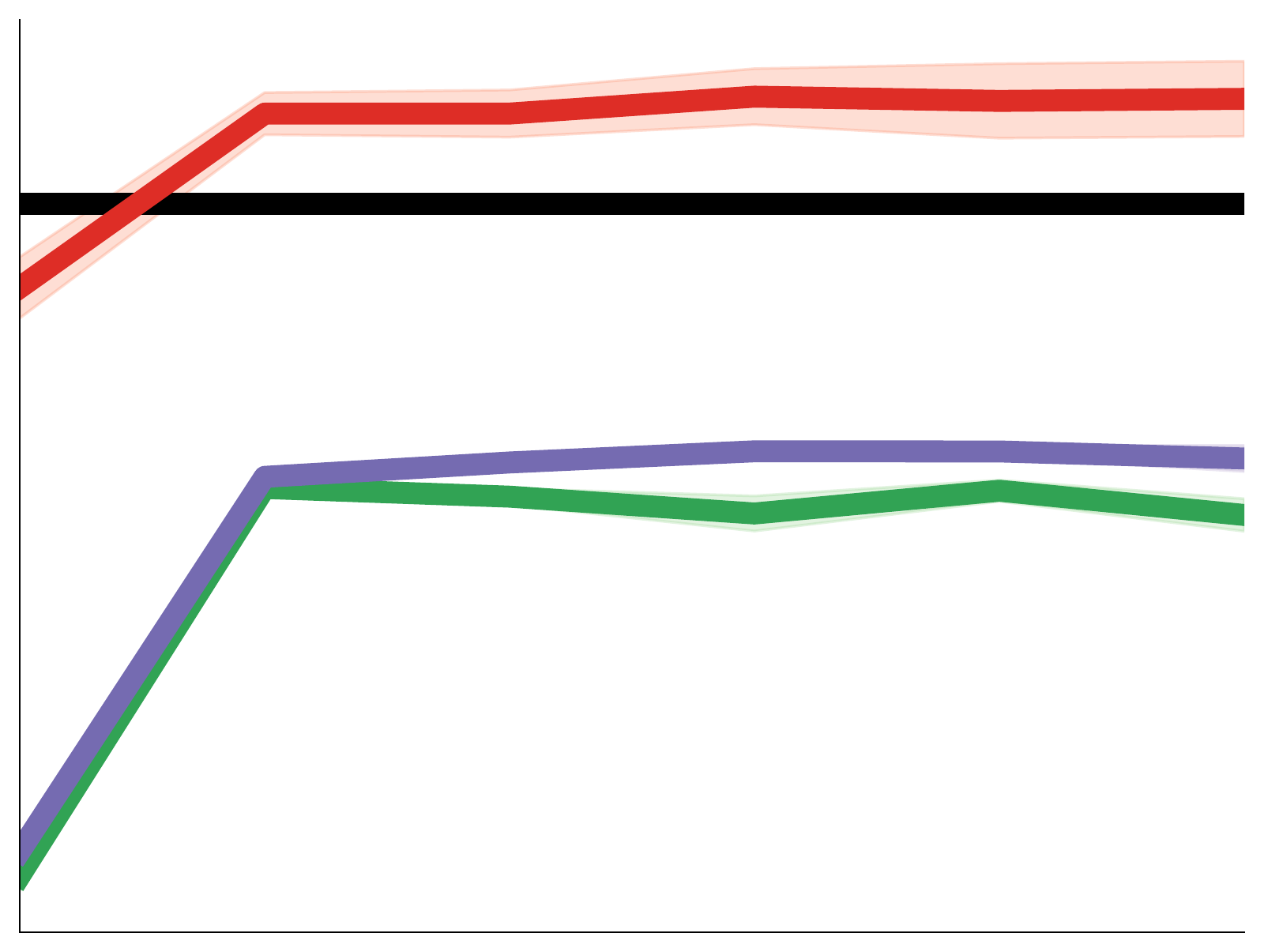}
    \caption{\envWamShelf}
    \label{fig:benchmark-wamshelf}
  \end{subfigure}
  ~
  \begin{subfigure}{0.15\linewidth}
    \centering
    \includegraphics[width=1\linewidth]{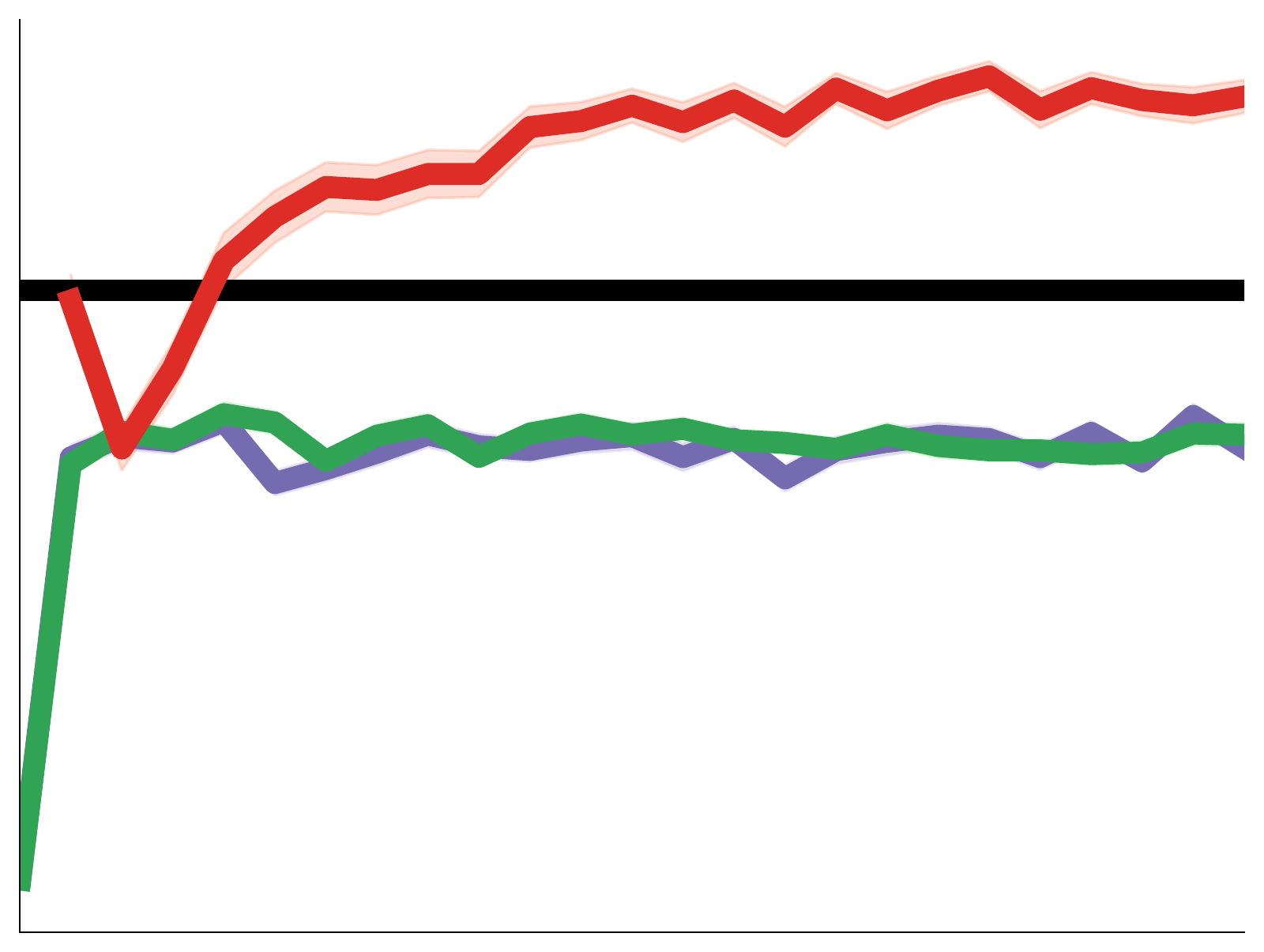}
    \caption{\envMaze}
    \label{fig:benchmark-maze}
  \end{subfigure}
  ~
  \begin{subfigure}{0.15\linewidth}
    \centering
    \includegraphics[width=1\linewidth]{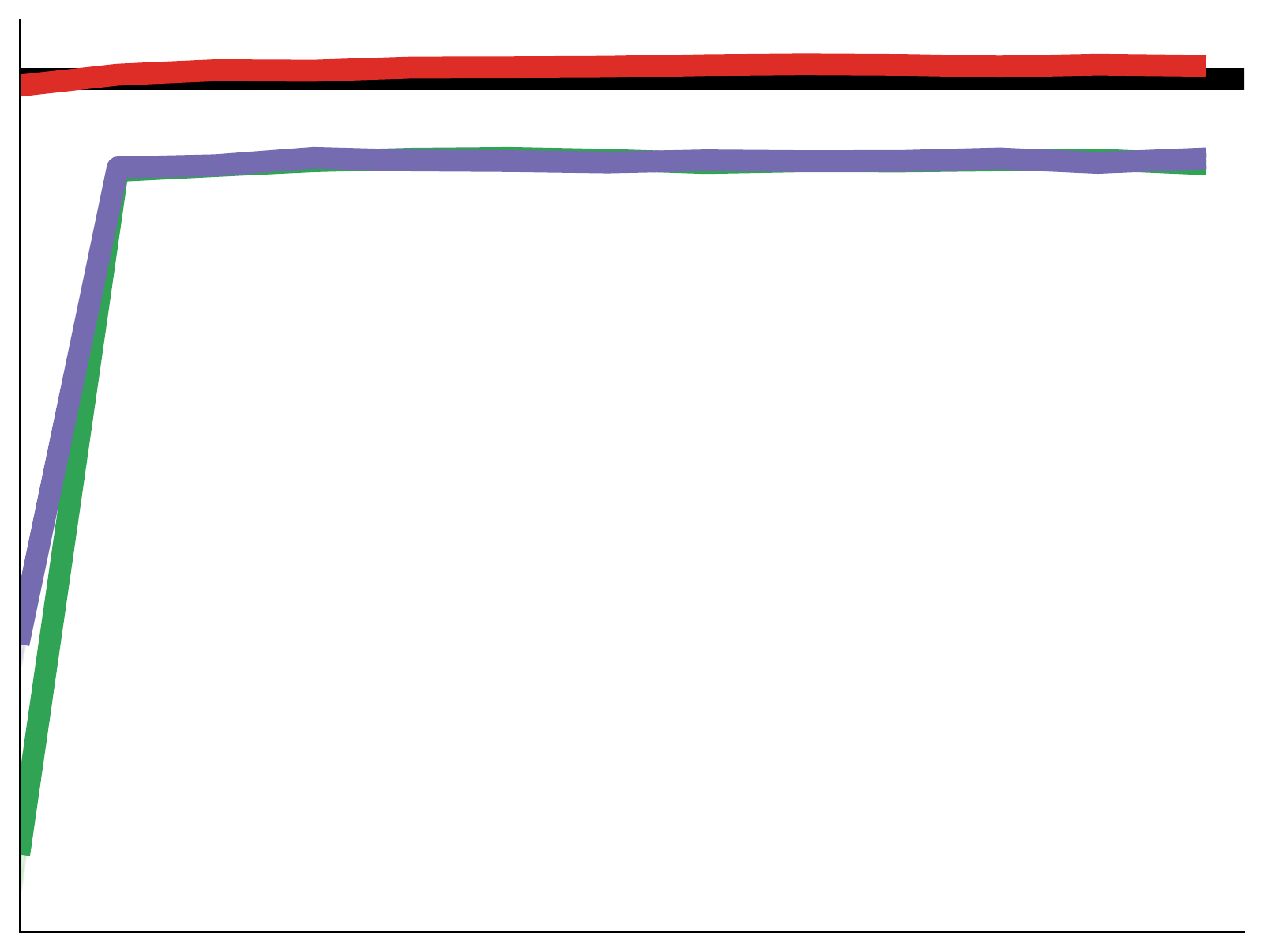}
    \caption{\envMazeTen}
  \end{subfigure}
  ~
  \begin{subfigure}{0.15\linewidth}
    \centering
    \includegraphics[width=1\linewidth]{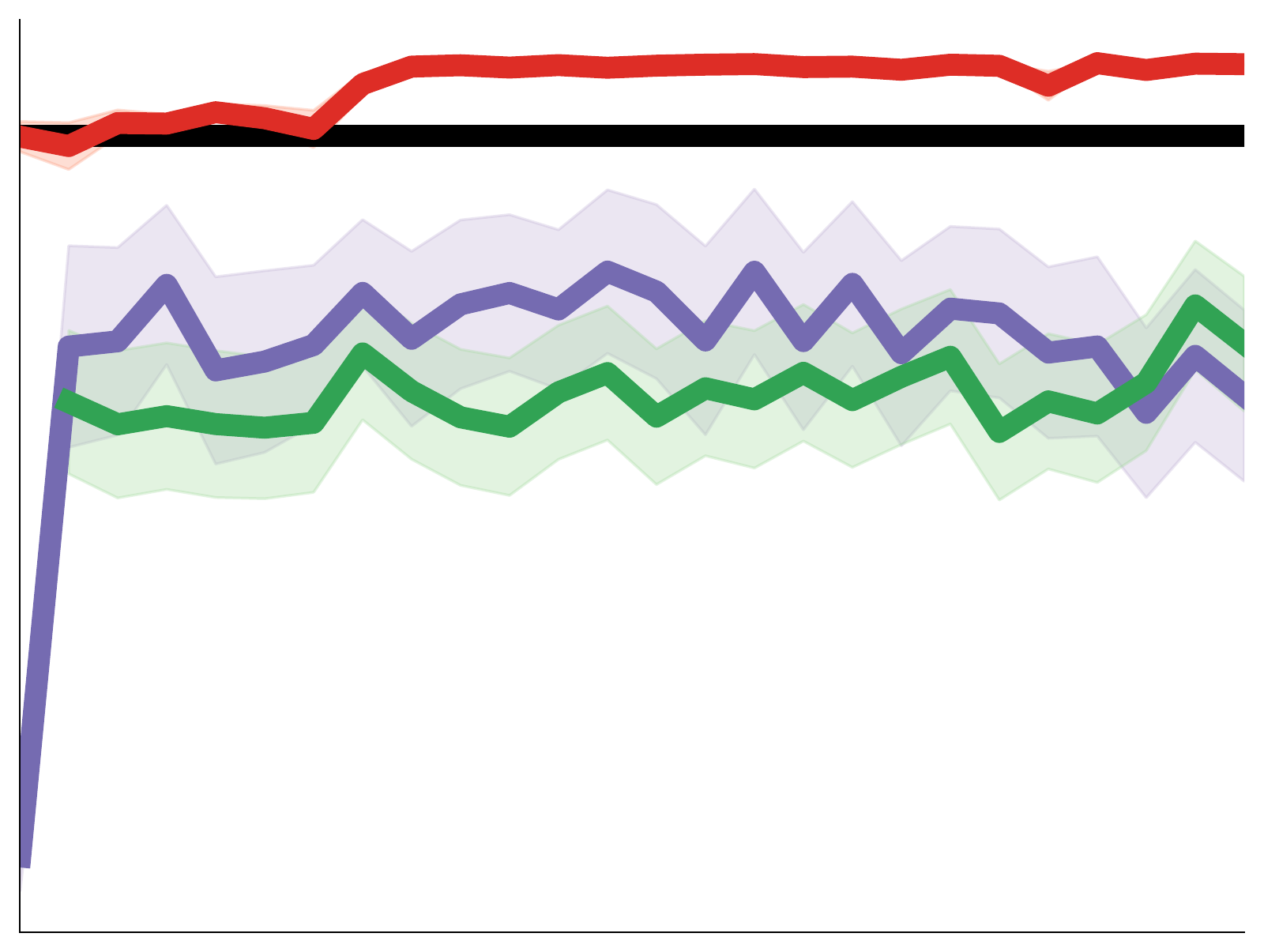}
    \caption{\envDoor}
  \end{subfigure}
  \caption{
    Training curves.
    \algName (red) dramatically outperforms agents that do not leverage expert knowledge (\algBPO in purple, \algUPMLE in green), and significantly improves the ensemble of experts (black).
    Axes have been scaled per environment to visualize the whole training curves for all algorithms.
  }
  \label{fig:benchmark}
\end{figure*}

\subsection{Environments}\label{ssec:envs}

Here we give a brief description of the problem environments.
\appsref{app:experiments-detail} contains implementation details.

\paragraph{Crowd Navigation}

Inspired by \citet{cai2019lets}, an autonomous agent must quickly navigate past a crowd of people without collisions.
Six people cross in front of the agent at fixed speeds, three from each side~(\Figref{fig:crosswalk-setup}).
Each person noisily walks toward its latent goal on the other side, which is sampled uniformly from a discrete set of destinations.
The agent observes each person's speed and position to estimate the belief distribution for each person's goal.
The belief for each person is drawn as a set of vectors in \Figref{fig:crosswalk-setup},
where length indicates speed and transparency indicates belief probability of each goal.
There is a single expert which uses model predictive control:
each walker is simulated toward a belief-weighted average goal position, and the expert selects cost-minimizing steering angle and acceleration.

\paragraph{Cartpole}

In this environment, the agent's goal is to keep the cartpole upright for as long as possible.
The latent parameters are cart mass and pole length (\figref{fig:cartpole-wam-setup}), uniformly sampled from $[0.5, 2.0] \mathrm{kg} \times [0.5, 2.0] \mathrm{m}$.
There is zero-mean Gaussian noise on the control.
The agent's estimator is a very coarse $3 \times 3$ discretization of the 2D continuous latent space, and the resulting belief is a categorical distribution over that grid.
In this environment, each expert is a Linear-Quadratic Regulator~(LQR) for the center of each grid square.
The ensemble recommendation used by \algName is simply the belief-weighted sum of experts, as described in \sref{ssec:ensemble}.

\paragraph{Object Localization}

In the {\bf \envWamShelf} environment, the agent must localize an object without colliding with the environment or object.
The continuous latent variable is the object's pose, which is anywhere on either shelf of the pantry (\figref{fig:wam-setup}).
The agent receives a noisy observation of the object's pose, which is very noisy when the agent does not invoke sensing.
Sensing can happen as the agent moves, and is less noisy the closer the end-effector is to the object.
The agent uses an Extended Kalman Filter to track the object's pose.
The ensemble is the MAP expert, as described in \sref{ssec:ensemble}.
It takes the MAP object pose and proposes a collision-free movement toward the object.

\paragraph{Latent Goal Mazes}

In the {\bf \envMaze} and {\bf \envMazeTen}, the agent must identify which latent goal is active.
At the beginning of each episode, the latent goal is set to one of four or ten goals.
The agent is rewarded highly for reaching the active goal and severely penalized for reaching an inactive goal.
Sensing can happen as the agent moves;
the agent receives a noisy measurement of the distance to the goal, with noise proportional to the true distance.
Each expert proposes an action (computed via motion planning) that navigates to the corresponding goal.
However, they are unaware of the penalty that corresponds to passing through an inactive goal.
The ensemble recommends the belief-weighted sum of the experts' suggestions.

\paragraph{Doors}

There are four possible doors to the next room of the {\bf \envDoor} environment.
At the beginning of each episode, each door is opened or closed with $0.5$ probability.
To check the doors, the agent can either sense or crash into them (which costs more than sensing).
Sensing is permitted while moving, and returns a noisy binary vector for all four doors with exponentially decreasing accuracy proportional to the distance to each door.
Crashing returns an accurate indicator of the door it crashed into.
Each expert navigates directly through the closest open door, and the ensemble recommends the belief-weighted sum of experts.

\begin{figure}
  \begin{subfigure}{1\linewidth}
    \centering
    \includegraphics[width=1\linewidth]{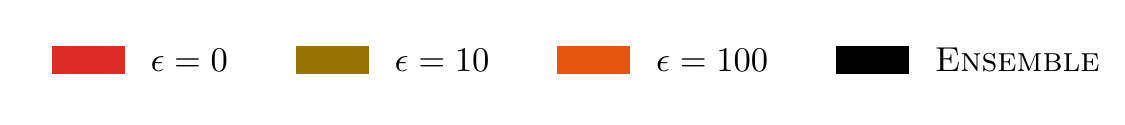}
  \end{subfigure}
  \\
  \begin{subfigure}{0.30\linewidth}
    \centering
    \includegraphics[width=1\linewidth]{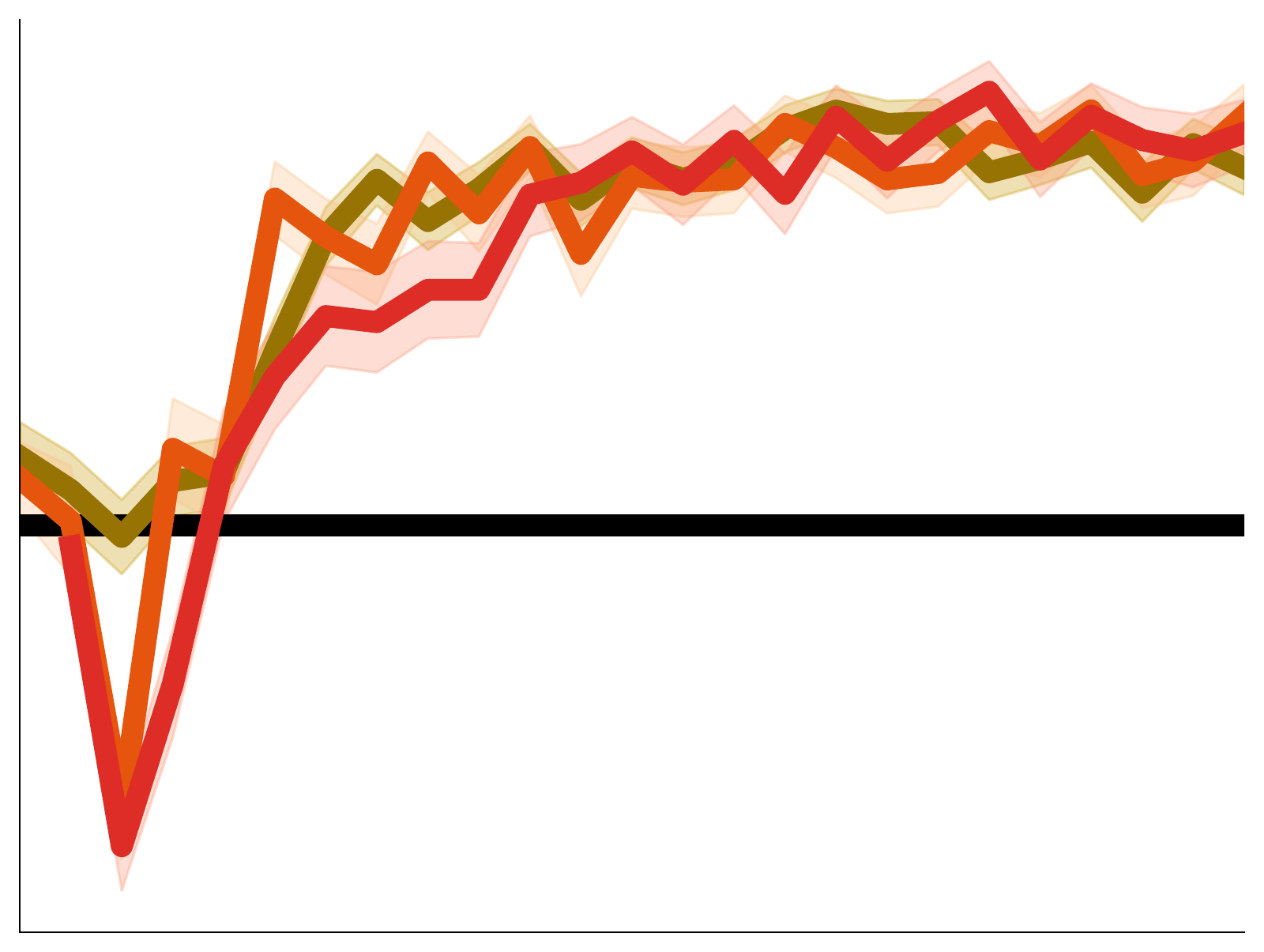}
    \caption{\envMaze}
  \end{subfigure}
  ~
  \begin{subfigure}{0.30\linewidth}
    \centering
    \includegraphics[width=1\linewidth]{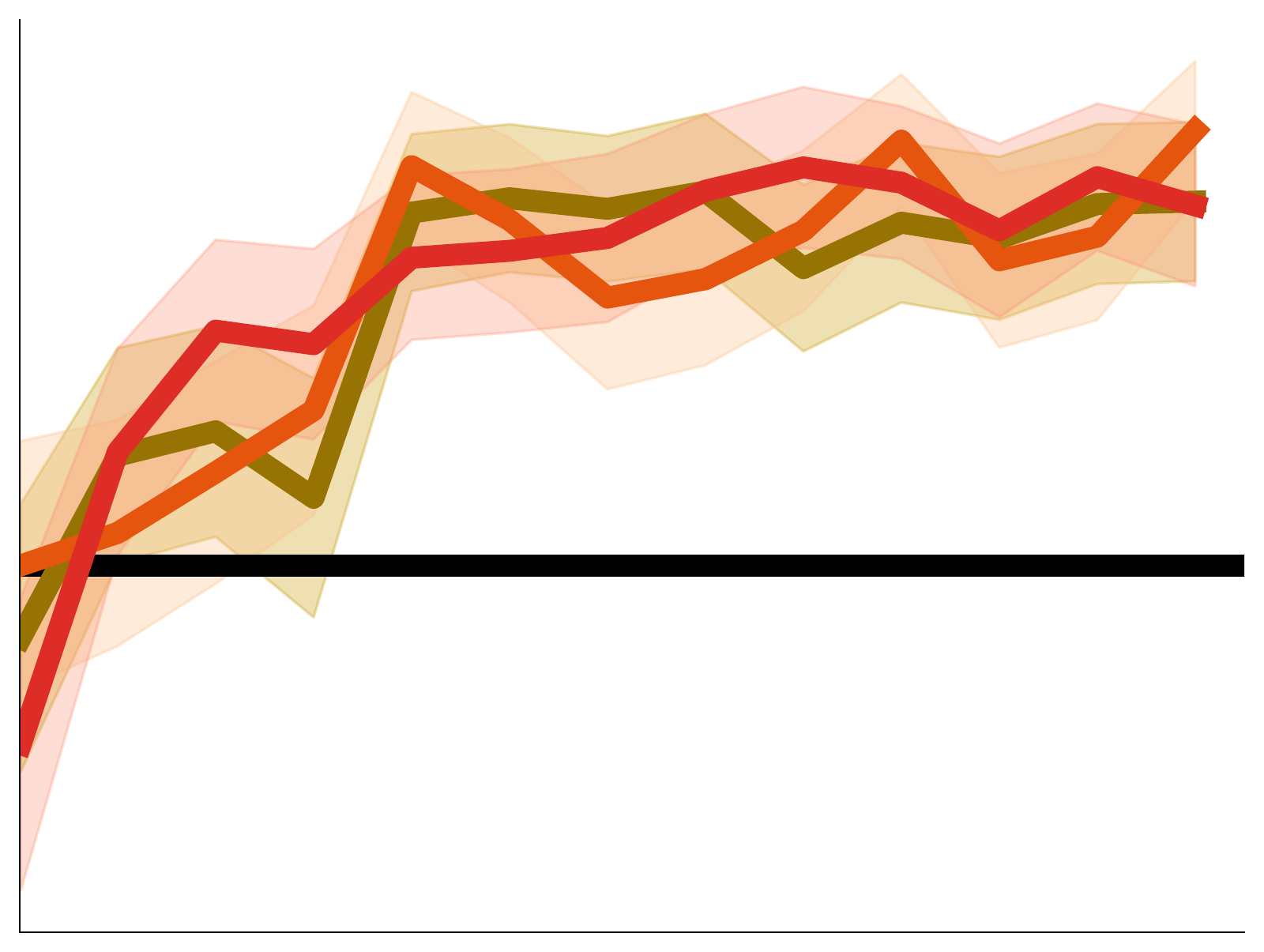}
    \caption{\envMazeTen}
  \end{subfigure}
  ~
  \begin{subfigure}{0.30\linewidth}
    \centering
    \includegraphics[width=1\linewidth]{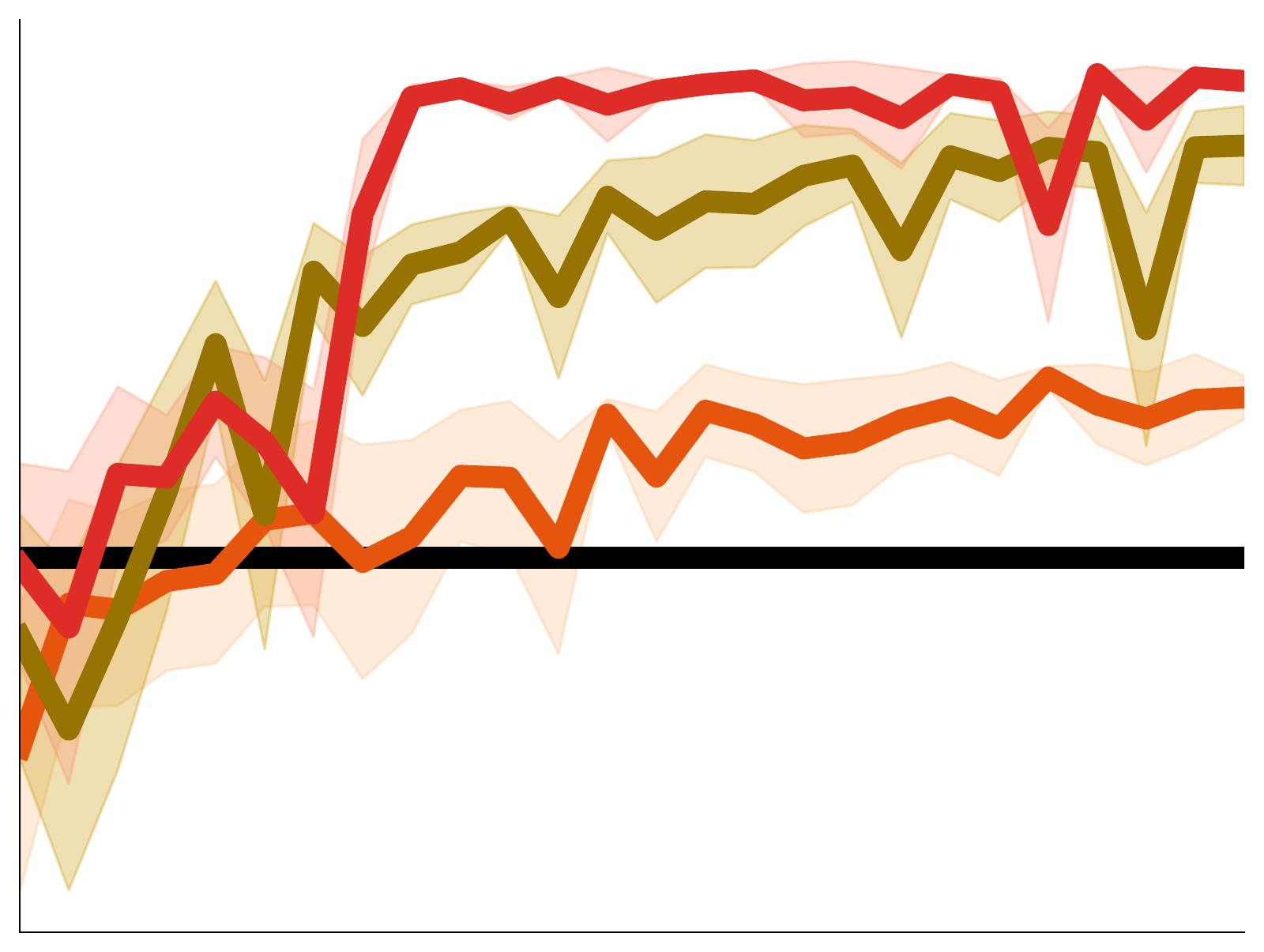}
    \caption{\envDoor}
  \end{subfigure}
  \caption{
    Ablation study on information-gathering reward (\Eqref{eq:info-gathering}).
    \algName is robust to information-gathering reward.
  }
  \label{fig:entropy_reward}
\end{figure}

  \begin{figure}
  \begin{subfigure}{0.3\linewidth}
    \centering
    \includegraphics[width=1\linewidth]{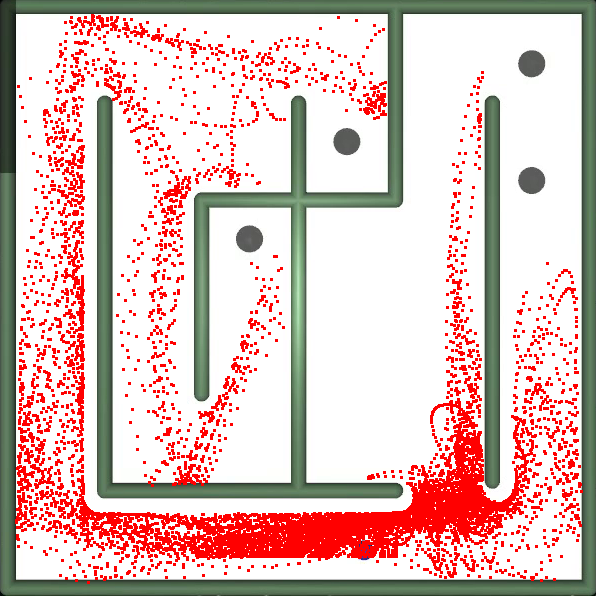}
    \caption{\envMaze}
  \end{subfigure}
  ~
  \begin{subfigure}{0.3\linewidth}
    \centering
    \includegraphics[width=1\linewidth]{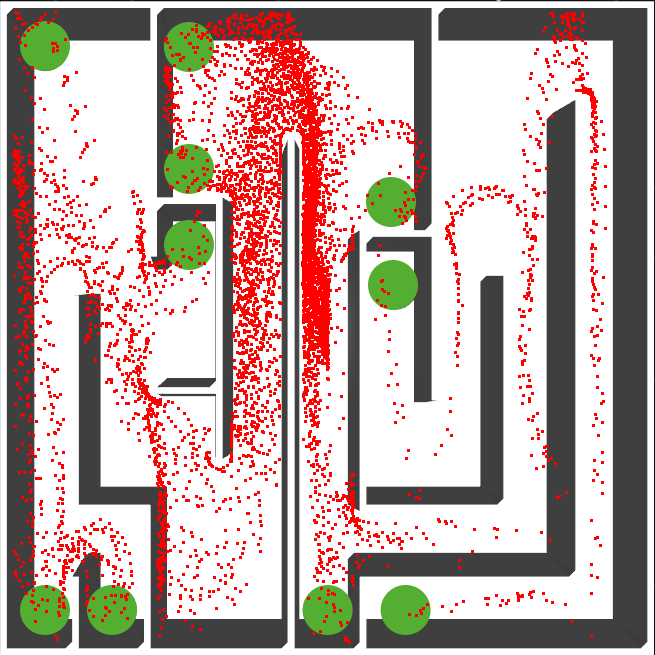}
    \caption{\envMazeTen}
  \end{subfigure}
  ~
  \begin{subfigure}{0.3\linewidth}
    \centering
    \includegraphics[width=1\linewidth]{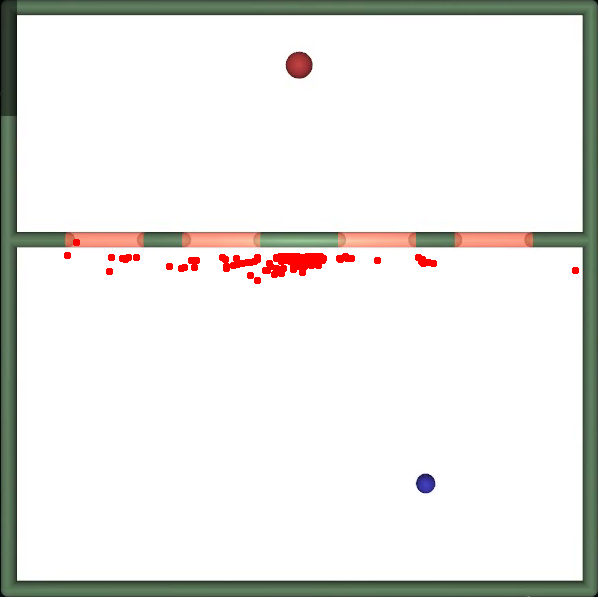}
    \caption{\envDoor}
  \end{subfigure}
  \vspace{0.5em}
  \caption{
    Sensing locations.
    In \envMaze and \envMazeTen, sensing is dense around the starting regions (the bottom row in \envMaze and center in \envMazeTen) and in areas where multiple latent goals are nearby.
    In \envDoor, \algName only senses when close to the doors, where the sensor is most accurate.
  }
  \label{fig:sensing}
\end{figure}

\begin{figure*}[t!]
  \centering
  \vspace{1em}
  \begin{subfigure}{1\linewidth}
    \centering
    \includegraphics[width=1\linewidth]{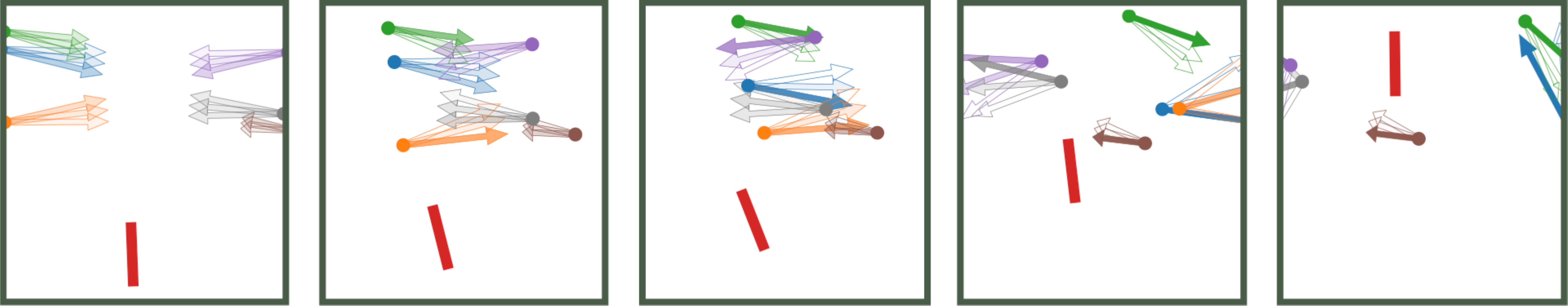}
    \vspace{-1.3em}
    \caption{{\bf\envCrosswalk}. Arrows are the directions to discrete latent goals. Each arrow's transparency indicate the posterior probability of the corresponding goal, and its length indicate the speed. The agent changes its direction as it forsees collision in its original plan.}
    \vspace{-0.5em}
    \label{fig:crosswalk-keyframes}
  \end{subfigure}
  \\
  \vspace{1em}
  \begin{subfigure}{1\linewidth}
    \centering
    \includegraphics[width=1\linewidth]{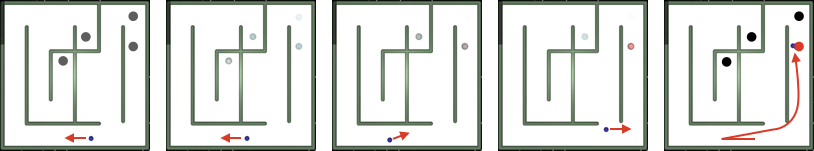}
    \\ \vspace{1em}
    \includegraphics[width=1\linewidth]{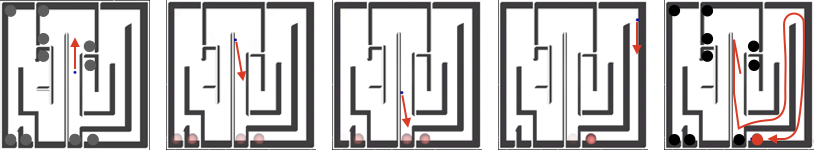}
    \caption{
      Latent goal mazes with four ({\bf \envMaze}) and ten ({\bf \envMazeTen}) possible goals.
      The agent senses as it navigates, changing its direction as goals are deemed less likely (more transparent).
      We have marked the true goal with red in the last frame for clarity.
    }
    \label{fig:maze-keyframes-both}
  \end{subfigure}
  \\
  \vspace{1em}
  \begin{subfigure}{1\linewidth}
    \centering
    \includegraphics[width=1\linewidth]{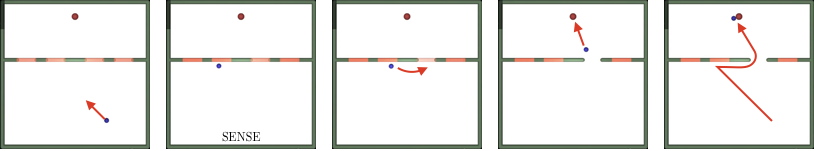}
    \caption{
      {\bf \envDoor}.
      The agent senses only when it is near the wall with doors, where sensing is most accurate.
      The transparency of the red bars indicates the posterior probability that the door is blocked.
      With sensing, the agent notices that the third door is likely to be open.
    }
    \label{fig:doors-keyframes}
  \end{subfigure}
  \caption{
    \algName policy keyframes. Best viewed in color.
  }
  \label{fig:experiments-highlight}
\end{figure*}

\subsection{BRPO Improves Ensemble, Outperforms Adaptive Methods}

We compare \algName to adaptive RL algorithms that consider the belief over latent states:
\algBPO~\citep{lee2018bayesian} and \algUPMLE, a modification to \citet{yu2017uposi} that augments the state with the most likely estimate from the Bayes filter\footnote{This was originally introduced in \citet{lee2018bayesian}.}.
Neither approach is able to incorporate experts.
We also compare with the ensemble of experts baselines.
For experiments which require explicit sensing actions~(\envWamShelf, \envMaze, \envMazeTen, \envDoor), the ensemble will not take any sensing actions (as discussed in \sref{sec:algorithm}), so we strengthen it by sensing with probability 0.5 at each timestep.
More sophisticated sensing strategies can be considered but require more task-specific knowledge to design;
see \appsref{app:ablation-ensemble} for more discussion.

\Figref{fig:benchmark} compares the training performance of all algorithms across the six environments.
In \sref{ssec:brpo-guarantee}, we proved monotonic improvement when optimizing an unconstrained objective;
the clipped surrogate PPO objective that \algName uses still yields improvement from the initial policy.
Note that \algName's initial policy does not exactly match the ensemble:
the random initialization for the residual policy network adds zero-mean noise around the ensemble policy.
This may result in an initial drop relative to the ensemble, as in \Figref{fig:benchmark-wamshelf} and \Figref{fig:benchmark-maze}.

On the wide variety of problems we have considered, \algName agents perform dramatically better than \algBPO and \algUPMLE agents.
\algBPO and \algUPMLE were unable to match the performance of \algName, except on the simple \envCartpole environment.
This seems to be due to the complexity of the latent MDPs, discussed further in \sref{ssec:qualitative-analysis}.
In fact, for \envMaze and \envMazeTen, we needed to modify the reward function to encourage information-gathering for \algBPO and \algUPMLE;
without such reward bonuses, they were unable to learn any meaningful behavior.
Even with the bonus, these agents only partially learn to solve the task.
We study the effect that such a reward bonus would have on \algName in \sref{ssec:reward-bonus}.
For the simpler \envCartpole environment, both \algBPO and \algUPMLE learned to perform optimally but required much more training time than \algName.

\subsection{Ablation Study: Information-Gathering Reward Bonuses}
\label{ssec:reward-bonus}

Because \algName maximizes the Bayesian Bellman equation~(\Eqref{eq:rl}), exploration is incorporated into its long-term objective.
As a result, auxiliary rewards to encourage exploration are unncessary.
However, existing work that does not explicitly consider the belief has suggested various auxiliary reward terms to encourage exploration, such as surprisal rewards~\citep{achiam2017surprise} or intrinsic rewards~\citep{pathak2017curiosity}.
To investigate whether such rewards benefit the \algName agent, we augment the reward function with the following auxiliary bonus from \cite{chen2016pomdp}:
\begin{equation}\label{eq:info-gathering}
\tilde{\reward}(\state, \belief, \action) = \reward(\state, \belief, \action) + \epsilon \cdot \mathbb{E}_{\belief'}\left[\|\belief - \belief'\|_1\right]
\end{equation}
where $\|\belief - \belief'\|_1 = \sum_{i=1}^k |\belief(\latent_i)  - \belief'(\latent_i)|$ rewards change in belief.

\Figref{fig:entropy_reward} summarizes the performance of \algName when training with $\epsilon=0, 10, 100$.
Too much emphasis on information-gathering causes the agent to over-explore and therefore underperform.
In \envDoor with $\epsilon=100$, we qualitatively observe that the agent crashes into the doors more often.
Crashing significantly changes the belief for that door;
the huge reward bonus outweighs the penalty of crashing from the environment.

We find that \algBPO and \algUPMLE are unable to learn without an exploration bonus on \envMaze, \envMazeTen, and \envDoor.
We used $\epsilon=1$ for \envMaze and \envDoor, and $\epsilon=100$ for \envMazeTen.
Upon qualitative analysis, we found that the bonus helps \algBPO and \algUPMLE learn to sense initially, but the algorithms are unable to make further progress.
We believe that this is because solving the latent mazes is challenging.

In addition to this study, we have performed two additional ablations on input features to the residual policy (\appsref{app:ablation-input}) and hand-tuned ensembles that are better at sensing (\appsref{app:ablation-ensemble}).
Including both the belief and ensemble recommendation as inputs to the residual policy produces faster learning.
\algName takes advantage of the stronger ensemble and continues to improve on that better baseline.

\subsection{Qualitative Behavior Analysis}
\label{ssec:qualitative-analysis}

\Figref{fig:experiments-highlight} shows some representative trajectories taken by \algName agents.
Across multiple environments~(\envCrosswalk, \envMaze, \envMazeTen), we see that \algName agent adapts to the evolving posterior.
As the posterior over latent goals updates, the agent shifts directions.
While this rerouting partly emerges from the ensemble policies as the posterior sharpens,
\algName's residual policy reduces uncertainty~(\envMaze, \envMazeTen) and pushes the agent to navigate faster, resulting in higher performance than the ensembles.

For \envMaze, \envMazeTen and \envDoor, we have visualized where the agent invokes explicit sensing~(\figref{fig:sensing}).
For \envMaze and \envMazeTen, the \algName agent learns to sense when goals must be distinguished, e.g. whenever the road diverges.
For \envDoor, it senses when that is most cost-effective: near the doors, where accuracy is highest.
This results in a rather interesting policy (\Figref{fig:doors-keyframes}).
The agent dashes to the wall, senses only once or twice, and drives through the closest open door.
The \algName agent avoids crashing in almost all scenarios.


\section{Discussion and Future Work}
\label{sec:discussion}

In the real world, robots must deal with uncertainty, either due to complex latent dynamics or task specifics.
Because uncertainty is an inherent part of these tasks, we can at best aim for optimality under uncertainty, i.e., Bayes optimality.
Existing Bayesian RL algorithms or POMDP solvers do not scale well to problems with complex continuous latent MDPs or a large set of possible MDPs.

Our algorithm, \algFullName, builds on an ensemble of experts by operating within the resulting residual belief MDP.
We prove that this strategy preserves guarantees, such as monotonic improvement, from the underlying policy optimization algorithm.
The scalability of policy gradient methods, combined with task-specific expertise, enables \algName to quickly solve a wide variety of complex problems, such as navigating through a crowd of pedestrians.
\algName improves on the original ensemble of experts and achieves much higher rewards than existing Bayesian RL algorithms by sensing more efficiently and acting more robustly.

Although out of scope for this work, a few key challenges remain.
First is an efficient construction of an ensemble of experts, which becomes particularly important for continuous latent spaces with infinitely many MDPs.
Infinitely many MDPs do not necessarily require infinite experts, as many may converge to similar policies.
An important future direction is subdividing the latent space and computing a qualitatively diverse set of policies~\citep{liu2016pac}.
Another challenge is developing an efficient Bayes filter, which is an active research area.
In certain occasions, the dynamics of the latent MDPs may not be accessible, which would require a learned Bayes filter.
Combined with a tractable, efficient Bayes filter and an efficiently computed set of experts, we believe that \algName will provide an even more scalable solution for BRL problems.

\bibliography{main}

\begin{thebibliography}{48}
\providecommand{\natexlab}[1]{#1}
\providecommand{\url}[1]{\texttt{#1}}
\expandafter\ifx\csname urlstyle\endcsname\relax
  \providecommand{\doi}[1]{doi: #1}\else
  \providecommand{\doi}{doi: \begingroup \urlstyle{rm}\Url}\fi

\bibitem[Achiam and Sastry(2017)]{achiam2017surprise}
Joshua Achiam and Shankar Sastry.
\newblock Surprise-based intrinsic motivation for deep reinforcement learning.
\newblock \emph{arXiv preprint arXiv:1703.01732}, 2017.

\bibitem[Baxter(1998)]{baxter1998theoretical}
Jonathan Baxter.
\newblock Theoretical models of learning to learn.
\newblock In \emph{Learning to learn}, pages 71--94. Springer, 1998.

\bibitem[Baxter(2000)]{baxter2000model}
Jonathan Baxter.
\newblock A model of inductive bias learning.
\newblock \emph{Journal of artificial intelligence research}, 12:\penalty0
  149--198, 2000.

\bibitem[Berkenkamp and Schoellig(2015)]{berkenkamp2015safe}
Felix Berkenkamp and Angela~P Schoellig.
\newblock Safe and robust learning control with gaussian processes.
\newblock In \emph{2015 European Control Conference (ECC)}, pages 2496--2501.
  IEEE, 2015.

\bibitem[Burda et~al.(2018)Burda, Edwards, Storkey, and Klimov]{Burda2018a}
Yuri Burda, Harrison Edwards, Amos Storkey, and Oleg Klimov.
\newblock {Exploration by random network distillation}.
\newblock \emph{arXiv preprint arXiv:1810.12894}, 2018.

\bibitem[Cai et~al.(2019)Cai, Luo, Saxena, Hsu, and Lee]{cai2019lets}
Panpan Cai, Yuanfu Luo, Aseem Saxena, David Hsu, and Wee~Sun Lee.
\newblock Lets-drive: Driving in a crowd by learning from tree search.
\newblock \emph{arXiv preprint arXiv:1905.12197}, 2019.

\bibitem[Chen et~al.(2016)Chen, Frazzoli, Hsu, and Lee]{chen2016pomdp}
Min Chen, Emilio Frazzoli, David Hsu, and Wee~Sun Lee.
\newblock {POMDP-lite} for robust robot planning under uncertainty.
\newblock In \emph{{IEEE} International Conference on Robotics and Automation},
  2016.

\bibitem[Choudhury et~al.(2018)Choudhury, Bhardwaj, Arora, Kapoor, Ranade,
  Scherer, and Dey]{choudhury2018data}
Sanjiban Choudhury, Mohak Bhardwaj, Sankalp Arora, Ashish Kapoor, Gireeja
  Ranade, Sebastian Scherer, and Debadeepta Dey.
\newblock Data-driven planning via imitation learning.
\newblock \emph{The International Journal of Robotics Research}, 37\penalty0
  (13-14):\penalty0 1632--1672, 2018.

\bibitem[Duan et~al.(2016)Duan, Schulman, Chen, Bartlett, Sutskever, and
  Abbeel]{duan2016rl}
Yan Duan, John Schulman, Xi~Chen, Peter~L Bartlett, Ilya Sutskever, and Pieter
  Abbeel.
\newblock Rl2: Fast reinforcement learning via slow reinforcement learning.
\newblock \emph{arXiv preprint arXiv:1611.02779}, 2016.

\bibitem[Finn et~al.(2017)Finn, Abbeel, and Levine]{finn2017model}
Chelsea Finn, Pieter Abbeel, and Sergey Levine.
\newblock Model-agnostic meta-learning for fast adaptation of deep networks.
\newblock In \emph{Proceedings of the 34th International Conference on Machine
  Learning-Volume 70}, pages 1126--1135. JMLR. org, 2017.

\bibitem[Finn et~al.(2018)Finn, Xu, and Levine]{finn2018promaml}
Chelsea Finn, Kelvin Xu, and Sergey Levine.
\newblock Probabilistic model-agnostic meta-learning.
\newblock \emph{arXiv preprint arXiv:1806.02817}, 2018.

\bibitem[Freund and Schapire(1999)]{freund1999short}
Yoav Freund and Robert Schapire.
\newblock A short introduction to boosting.
\newblock \emph{Journal-Japanese Society For Artificial Intelligence},
  14\penalty0 (771-780):\penalty0 1612, 1999.

\bibitem[Ghavamzadeh et~al.(2015)Ghavamzadeh, Mannor, Pineau, Tamar,
  et~al.]{ghavamzadeh2015bayesian}
Mohammad Ghavamzadeh, Shie Mannor, Joelle Pineau, Aviv Tamar, et~al.
\newblock {Bayesian} reinforcement learning: A survey.
\newblock \emph{Foundations and Trends{\textregistered} in Machine Learning},
  8\penalty0 (5-6):\penalty0 359--483, 2015.

\bibitem[Gimelfarb et~al.(2018)Gimelfarb, Sanner, and
  Lee]{gimelfarb2018reinforcement}
Michael Gimelfarb, Scott Sanner, and Chi-Guhn Lee.
\newblock Reinforcement learning with multiple experts: A bayesian model
  combination approach.
\newblock In \emph{Advances in Neural Information Processing Systems}, pages
  9528--9538, 2018.

\bibitem[Grant et~al.(2018)Grant, Finn, Levine, Darrell, and
  Griffiths]{grant2018recasting}
Erin Grant, Chelsea Finn, Sergey Levine, Trevor Darrell, and Thomas Griffiths.
\newblock Recasting gradient-based meta-learning as hierarchical bayes.
\newblock \emph{arXiv preprint arXiv:1801.08930}, 2018.

\bibitem[Guez et~al.(2012)Guez, Silver, and Dayan]{guez2012efficient}
Arthur Guez, David Silver, and Peter Dayan.
\newblock Efficient {Bayes}-adaptive reinforcement learning using sample-based
  search.
\newblock In \emph{Advances in Neural Information Processing Systems}, 2012.

\bibitem[Gupta et~al.(2018)Gupta, Mendonca, Liu, Abbeel, and
  Levine]{gupta2018meta}
Abhishek Gupta, Russell Mendonca, YuXuan Liu, Pieter Abbeel, and Sergey Levine.
\newblock Meta-reinforcement learning of structured exploration strategies.
\newblock In \emph{Advances in Neural Information Processing Systems}, pages
  5302--5311, 2018.

\bibitem[Hochreiter and Schmidhuber(1997)]{hochreiter1997long}
Sepp Hochreiter and J{\"u}rgen Schmidhuber.
\newblock Long short-term memory.
\newblock \emph{Neural computation}, 9\penalty0 (8):\penalty0 1735--1780, 1997.

\bibitem[Houthooft et~al.(2016)Houthooft, Chen, Duan, Schulman, De~Turck, and
  Abbeel]{houthooft2016vime}
Rein Houthooft, Xi~Chen, Yan Duan, John Schulman, Filip De~Turck, and Pieter
  Abbeel.
\newblock Vime: Variational information maximizing exploration.
\newblock In \emph{Advances in Neural Information Processing Systems}, pages
  1109--1117, 2016.

\bibitem[Hsu et~al.(2008)Hsu, Lee, and Rong]{hsu2008makes}
David Hsu, Wee~S Lee, and Nan Rong.
\newblock What makes some pomdp problems easy to approximate?
\newblock In \emph{Advances in neural information processing systems}, pages
  689--696, 2008.

\bibitem[Johannink et~al.(2019)Johannink, Bahl, Nair, Luo, Kumar, Loskyll,
  Ojea, Solowjow, and Levine]{johannink2019residual}
Tobias Johannink, Shikhar Bahl, Ashvin Nair, Jianlan Luo, Avinash Kumar,
  Matthias Loskyll, Juan~Aparicio Ojea, Eugen Solowjow, and Sergey Levine.
\newblock Residual reinforcement learning for robot control.
\newblock In \emph{2019 International Conference on Robotics and Automation
  (ICRA)}, pages 6023--6029. IEEE, 2019.

\bibitem[Kahn et~al.(2017)Kahn, Zhang, Levine, and Abbeel]{kahn2017plato}
Gregory Kahn, Tianhao Zhang, Sergey Levine, and Pieter Abbeel.
\newblock Plato: Policy learning using adaptive trajectory optimization.
\newblock In \emph{{IEEE} International Conference on Robotics and Automation},
  pages 3342--3349. IEEE, 2017.

\bibitem[Kolter and Ng(2009)]{kolter2009near}
Zico Kolter and Andrew Ng.
\newblock Near-{Bayesian} exploration in polynomial time.
\newblock In \emph{International Conference on Machine Learning}, 2009.

\bibitem[Kurniawati et~al.(2008)Kurniawati, Hsu, and Lee]{kurniawati2008sarsop}
Hanna Kurniawati, David Hsu, and Wee~Sun Lee.
\newblock {SARSOP}: Efficient point-based {POMDP} planning by approximating
  optimally reachable belief spaces.
\newblock In \emph{Robotics: Science and Systems}, 2008.

\bibitem[Lee et~al.(2019)Lee, Hou, Mandalika, Lee, Choudhury, and
  Srinivasa]{lee2018bayesian}
Gilwoo Lee, Brian Hou, Aditya Mandalika, Jeongseok Lee, Sanjiban Choudhury, and
  Siddhartha~S. Srinivasa.
\newblock Bayesian policy optimization for model uncertainty.
\newblock In \emph{International Conference on Learning Representations}, 2019.

\bibitem[Liu et~al.(2016)Liu, Guo, and Brunskill]{liu2016pac}
Yao Liu, Zhaohan Guo, and Emma Brunskill.
\newblock Pac continuous state online multitask reinforcement learning with
  identification.
\newblock In \emph{Proceedings of the 2016 International Conference on
  Autonomous Agents \& Multiagent Systems}, pages 438--446. International
  Foundation for Autonomous Agents and Multiagent Systems, 2016.

\bibitem[Mendonca et~al.(2019)Mendonca, Gupta, Kralev, Abbeel, Levine, and
  Finn]{mendonca2019guided}
Russell Mendonca, Abhishek Gupta, Rosen Kralev, Pieter Abbeel, Sergey Levine,
  and Chelsea Finn.
\newblock Guided meta-policy search.
\newblock \emph{arXiv preprint arXiv:1904.00956}, 2019.

\bibitem[Ortega et~al.(2019)Ortega, Wang, Rowland, Genewein, Kurth{-}Nelson,
  Pascanu, Heess, Veness, Pritzel, Sprechmann, Jayakumar, McGrath, Miller,
  Azar, Osband, Rabinowitz, Gy{\"{o}}rgy, Chiappa, Osindero, Teh, van Hasselt,
  de~Freitas, Botvinick, and Legg]{ortega2019meta}
Pedro~A. Ortega, Jane~X. Wang, Mark Rowland, Tim Genewein, Zeb Kurth{-}Nelson,
  Razvan Pascanu, Nicolas Heess, Joel Veness, Alexander Pritzel, Pablo
  Sprechmann, Siddhant~M. Jayakumar, Tom McGrath, Kevin Miller,
  Mohammad~Gheshlaghi Azar, Ian Osband, Neil~C. Rabinowitz, Andr{\'{a}}s
  Gy{\"{o}}rgy, Silvia Chiappa, Simon Osindero, Yee~Whye Teh, Hado van Hasselt,
  Nando de~Freitas, Matthew Botvinick, and Shane Legg.
\newblock Meta-learning of sequential strategies.
\newblock \emph{arXiv preprint arXiv:1905.03030}, 2019.

\bibitem[Osband et~al.(2013)Osband, Russo, and Van~Roy]{osband2013psrl}
Ian Osband, Daniel Russo, and Benjamin Van~Roy.
\newblock (more) efficient reinforcement learning via posterior sampling.
\newblock In \emph{Advances in Neural Information Processing Systems}, 2013.

\bibitem[Osband et~al.(2019)Osband, Roy, Russo, and Wen]{JMLR:v20:18-339}
Ian Osband, Benjamin~Van Roy, Daniel~J. Russo, and Zheng Wen.
\newblock Deep exploration via randomized value functions.
\newblock \emph{Journal of Machine Learning Research}, 20\penalty0
  (124):\penalty0 1--62, 2019.
\newblock URL \url{http://jmlr.org/papers/v20/18-339.html}.

\bibitem[Ostafew et~al.(2014)Ostafew, Schoellig, and
  Barfoot]{ostafew2014learning}
Chris~J Ostafew, Angela~P Schoellig, and Timothy~D Barfoot.
\newblock Learning-based nonlinear model predictive control to improve
  vision-based mobile robot path-tracking in challenging outdoor environments.
\newblock In \emph{{IEEE} International Conference on Robotics and Automation},
  pages 4029--4036. IEEE, 2014.

\bibitem[Ostafew et~al.(2015)Ostafew, Schoellig, and
  Barfoot]{ostafew2015conservative}
Chris~J Ostafew, Angela~P Schoellig, and Timothy~D Barfoot.
\newblock Conservative to confident: treating uncertainty robustly within
  learning-based control.
\newblock In \emph{{IEEE} International Conference on Robotics and Automation},
  pages 421--427. IEEE, 2015.

\bibitem[Pathak et~al.(2017)Pathak, Agrawal, Efros, and
  Darrell]{pathak2017curiosity}
Deepak Pathak, Pulkit Agrawal, Alexei~A Efros, and Trevor Darrell.
\newblock Curiosity-driven exploration by self-supervised prediction.
\newblock In \emph{Proceedings of the IEEE Conference on Computer Vision and
  Pattern Recognition Workshops}, pages 16--17, 2017.

\bibitem[Peng et~al.(2018)Peng, Andrychowicz, Zaremba, and Abbeel]{peng2018sim}
Xue~Bin Peng, Marcin Andrychowicz, Wojciech Zaremba, and Pieter Abbeel.
\newblock Sim-to-real transfer of robotic control with dynamics randomization.
\newblock In \emph{{IEEE} International Conference on Robotics and Automation},
  2018.

\bibitem[Pineau et~al.(2003)Pineau, Gordon, Thrun, et~al.]{pineau2003point}
Joelle Pineau, Geoff Gordon, Sebastian Thrun, et~al.
\newblock Point-based value iteration: An anytime algorithm for {POMDPs}.
\newblock In \emph{International Joint Conference on Artificial Intelligence},
  2003.

\bibitem[Rabinowitz(2019)]{rabinowitz2019meta}
Neil~C. Rabinowitz.
\newblock Meta-learners' learning dynamics are unlike learners'.
\newblock \emph{arXiv preprint arXiv:1905.01320}, 2019.

\bibitem[Rajeswaran et~al.(2017)Rajeswaran, Ghotra, Ravindran, and
  Levine]{rajeswaran2016epopt}
Aravind Rajeswaran, Sarvjeet Ghotra, Balaraman Ravindran, and Sergey Levine.
\newblock {EPOpt}: Learning robust neural network policies using model
  ensembles.
\newblock In \emph{International Conference on Learning Representations}, 2017.

\bibitem[Rakelly et~al.(2019)Rakelly, Zhou, Quillen, Finn, and
  Levine]{rakelly2019efficient}
Kate Rakelly, Aurick Zhou, Deirdre Quillen, Chelsea Finn, and Sergey Levine.
\newblock Efficient off-policy meta-reinforcement learning via probabilistic
  context variables.
\newblock \emph{arXiv preprint arXiv:1903.08254}, 2019.

\bibitem[Schulman et~al.(2015)Schulman, Levine, Abbeel, Jordan, and
  Moritz]{schulman2015trpo}
John Schulman, Sergey Levine, Pieter Abbeel, Michael Jordan, and Philipp
  Moritz.
\newblock Trust region policy optimization.
\newblock In \emph{International Conference on Machine Learning}, 2015.

\bibitem[Schulman et~al.(2017)Schulman, Wolski, Dhariwal, Radford, and
  Klimov]{schulman2017proximal}
John Schulman, Filip Wolski, Prafulla Dhariwal, Alec Radford, and Oleg Klimov.
\newblock Proximal policy optimization algorithms.
\newblock \emph{arXiv preprint arXiv:1707.06347}, 2017.

\bibitem[Shani et~al.(2013)Shani, Pineau, and Kaplow]{shani2013survey}
Guy Shani, Joelle Pineau, and Robert Kaplow.
\newblock A survey of point-based {POMDP} solvers.
\newblock \emph{Journal on Autonomous Agents and Multiagent Systems},
  27\penalty0 (1):\penalty0 1--51, 2013.

\bibitem[Silver and Veness(2010)]{silver2010monte}
David Silver and Joel Veness.
\newblock Monte-carlo planning in large {POMDPs}.
\newblock In \emph{Advances in Neural Information Processing Systems}, 2010.

\bibitem[Silver et~al.(2018)Silver, Allen, Tenenbaum, and
  Kaelbling]{silver2018residual}
Tom Silver, Kelsey Allen, Josh Tenenbaum, and Leslie Kaelbling.
\newblock Residual policy learning.
\newblock \emph{arXiv preprint arXiv:1812.06298}, 2018.

\bibitem[Stadie et~al.(2018)Stadie, Yang, Houthooft, Chen, Duan, Wu, Abbeel,
  and Sutskever]{stadie2018meta}
Bradly~C. Stadie, Ge~Yang, Rein Houthooft, Xi~Chen, Yan Duan, Yuhuai Wu, Pieter
  Abbeel, and Ilya Sutskever.
\newblock Some considerations on learning to explore via meta-reinforcement
  learning.
\newblock \emph{arXiv preprint arXiv:1803.01118}, 2018.

\bibitem[Sunberg and Kochenderfer(2018)]{sunberg2018pomcpow}
Zachary Sunberg and Mykel Kochenderfer.
\newblock Online algorithms for {POMDPs} with continuous state, action, and
  observation spaces.
\newblock In \emph{International Conference on Automated Planning and
  Scheduling}, 2018.

\bibitem[Tobin et~al.(2017)Tobin, Fong, Ray, Schneider, Zaremba, and
  Abbeel]{tobin2017domain}
Josh Tobin, Rachel Fong, Alex Ray, Jonas Schneider, Wojciech Zaremba, and
  Pieter Abbeel.
\newblock Domain randomization for transferring deep neural networks from
  simulation to the real world.
\newblock In \emph{{IEEE/RSJ} International Conference on Intelligent Robots
  and Systems}, 2017.

\bibitem[Yoon et~al.(2018)Yoon, Kim, Dia, Kim, Bengio, and
  Ahn]{yoon2018bayesian}
Jaesik Yoon, Taesup Kim, Ousmane Dia, Sungwoong Kim, Yoshua Bengio, and Sungjin
  Ahn.
\newblock Bayesian model-agnostic meta-learning.
\newblock In \emph{Advances in Neural Information Processing Systems}, pages
  7332--7342, 2018.

\bibitem[Yu et~al.(2017)Yu, Tan, Liu, and Turk]{yu2017uposi}
Wenhao Yu, Jie Tan, C.~Karen Liu, and Greg Turk.
\newblock Preparing for the unknown: Learning a universal policy with online
  system identification.
\newblock In \emph{Robotics: Science and Systems}, 2017.

\end{thebibliography}
\bibliographystyle{plainnat}

\clearpage

\appendix{}

\subsection{Experimental Environments}\label{app:experiments-detail}

\paragraph{Crowd Navigation}

At the beginning of the episode, initial pedestrian positions are sampled uniformly along the left and right sides of the environment.
Speeds are sampled uniformly between 0.1 and 1.0 m/s.
The agent observes each person's speed and position to estimate the goal distribution.

The agent starts at the bottom of the environment, with initial speed sampled uniformly from 0 to 0.4 m/s.
The agent controls acceleration and steering angle, bounded between $\pm0.12$ m/$s^2$ and $\pm0.1$ rad.
Pedestrians are modeled as a 1m diameter circle.
The agent is modeled as a rectangular vehicle of $0.5$~m width and $2$~m length.
A collision results in a terminal cost of $100\cdot(2v)^2+0.5$.
Successfully reaching the top of the environment produces terminal reward of $250$, while navigating to the left or right side results in terminal cost of $1000$.
A per timestep penalty of $0.1$ encourages the agent to complete the episode quickly.

\paragraph{Cartpole}

The cartpole initializes with small initial velocity around the upright position.
The environment terminates when the pole is more than $1.2$~rad away from the vertical upright position or the cart is $4.0$~m away from the center.
The agent is rewarded by 1 for every step the cartpole survives.
The environment has finite horizon of 500 steps.

\paragraph{Object Localization}

The agent can control the end-effector in the $(x, y, z)$ directions.
The goal is to move the hand to the object without colliding with the environment or object.
The agent observes the top and bottom shelf poses, end-effector pose, arm configuration, and the noise scale.
The noise scale is the standard deviation of the Gaussian noise on the agent's observation of the object's pose.
Without sensing, the noise is very large: $w \sim \mathcal{N}(0, 5.0^2)$ where the width of the shelf is only $0.35$ m,
When sensing is invoked, the noise is reduced to $w \sim\mathcal{N}(0, d^2)$ where $d$ is the distance between the object and the end-effector.

\paragraph{Latent Goal Mazes}

The agent observes its current position, velocity, and distance to all latent goals.
If sensing is invoked, it also observes the noisy distance to the goal.
In addition, the agent observes the categorical belief distribution over the latent goals.
In \envMaze, reaching the active goal provides a terminal reward of $500$, while reaching an incorrect goal gives a penalty of $500$.
The task ends when the agent receives either the terminal reward or penalty, or after 500 timesteps.
In \envMazeTen, the agent receives a penalty of $50$ and continues to explore after reaching an incorrect goal.

\paragraph{Doors}
To check the doors, the agent can either sense ($-1$) or crash into them ($-10$).
At every step, the agent observes its position, velocity, distance to goal, and whether it crashed or passed through a door.
In addition, the agent observes the categorical distribution over the $2^4 = 16$ possible door configurations (from the Bayes filter) and the ensemble's recommendation.
The agent receives a terminal reward of 100 if it reaches the goal within 300 timesteps.

\subsection{Bayesian Reinforcement Learning and Posterior Sampling}\label{ssec:psrl}
Posterior Sampling Reinforcement Learning (PSRL)~\citep{osband2013psrl} is an online RL algorithm that maintains a posterior over latent MDP parameters $\phi$.
However, the problem setting it considers and how it uses this posterior are quite different than what we consider in this paper.

In this work, we are focused on one-shot scenarios where the agent can only interact with the test MDP for a single episode; latent parameters are resampled for each episode.
The PSRL regret analysis assumes MDPs with finite horizons and repeated episodes with the same test MDP, i.e. the latent parameters are fixed for all episodes.

Before each episode, PSRL samples an MDP from its posterior over MDPs, computes the optimal policy for the sampled MDP, and executes it on the fixed test MDP.
Its posterior is updated after each episode, concentrating the distribution around the true latent parameters.
During this exploration period, it can perform arbitrarily poorly.
Furthermore, sampling a latent MDP from the posterior determinizes the parameters;
as a result, there is no uncertainty in the sampled MDP, and the resulting optimal policies that are executed will never take sensing actions.

\paragraph{The Gap Between Bayes Optimality and Posterior Sampling in one-shot setting}

We present a toy problem to highlight the distinction between them.

\begin{figure}[t!]
\centering
\begin{subfigure}[b]{1\linewidth}
\centering
\includegraphics[width=1\linewidth]{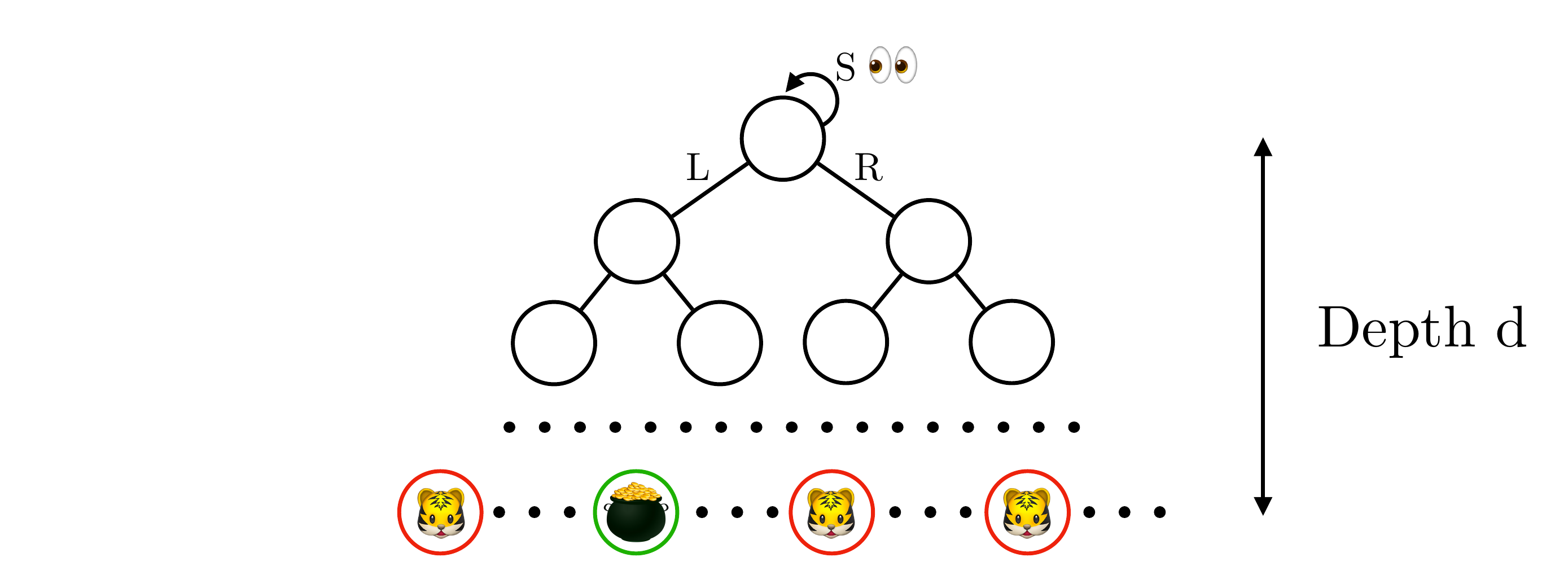}
\end{subfigure}
\caption{
A tree-like MDP that highlights the distinction between BRL and PSRL.
}
\label{fig:tree_mdp}
\end{figure}

Consider a deterministic tree-like MDP (\figref{fig:tree_mdp}).
Reward is received only at the terminal leaf states:
one leaf contains a pot of gold ($R=100$) and all others contain a dangerous tiger ($R=-10$).
All non-leaf states have two actions, go left (L) and go right (R).
The start state additionally has a sense action (S), which is costly ($R=-0.1$) but reveals the exact location of the pot of gold.
Both algorithms are initialized with a uniform prior over the $N = 2^d$ possible MDPs (one for each possible location of the pot of gold).

To contrast the performance of the Bayes-optimal policy and posterior sampling, we consider the multi-episode setting where the agent repeatedly interacts with the same MDP.
The MDP is sampled once from the uniform prior, and agents interact with it for $T$ episodes.
This is the setting typically considered by posterior sampling (PSRL)~\citep{osband2013psrl}.

Before each episode, PSRL samples an MDP from its posterior over MDPs, computes the optimal policy, and executes it.
After each episode, it updates the posterior and repeats.
Sampling from the posterior determinizes the underlying latent parameter.
As a result, PSRL will never produce sensing actions to reduce uncertainty about that parameter because the sampled MDP has no uncertainty.
More concretely, the optimal policy for each tree MDP is to navigate directly to the gold \textit{without sensing};
PSRL will never take the sense action.
Thus, PSRL makes an average of $\frac{N-1}{2}$ mistakes before sampling the correct pot of gold location and the cumulative reward over $T$ episodes is
\begin{equation}
-10\underbrace{\left(\textstyle\frac{N-1}{2}\right)}_{\text{mistakes}} + 100\underbrace{\left(T - \textstyle\frac{N-1}{2}\right)}_{\text{pot of gold}}
\end{equation}
In the first episode, the Bayes-optimal first action is to sense.
All subsequent actions in this first episode navigate toward the pot of gold, for an episode reward of $-0.1 + 100$.
In the subsequent $T-1$ episodes, the Bayes-optimal policy navigates directly toward the goal without needing to sense, for a cumulative reward of $100T - 0.1$.
The performance gap between the Bayes-optimal policy and posterior sampling grows exponentially with depth of the tree $d$.

Practically, a na\"ive policy gradient algorithm (like BPO) would struggle to learn the Bayes-optimal policy: it would need to learn to both sense and navigate the tree to the sensed goal.
\algName can take advantage of the set of experts, in which each navigate to their designated leaf.
During training, the \algName agent only needs to learn to balance sensing with navigation.

\subsection{Maximum A Posterior as ensemble of experts}
\label{ssec:appendix-map-expert}
One choice for the ensemble policy $\policy_e$ is to select the maximum a posteriori (MAP) action, $a_{\rm MAP} = \argmax_a \sum_{i=1}^k \belief(\phi_i) \policy_i(a | s)$.
However, computing the MAP estimate may require optimizing a non-convex function, e.g., when the distribution is multimodal.
We can instead maximize the lower bound using Jensen's inequality.
\begin{equation}
\label{eq:expected_log}
\log \sum_{i=1}^k b(\phi_i) \pi_i(a | s) \geq \sum_{i=1}^k b(\phi_i) \log \pi_i(a | s)
\end{equation}
This is much easier to solve, especially if $\log \policy_i(a | s)$ is convex.
If each $\pi_i(a | s)$ is a Gaussian with mean $\mu_i$ and covariance $\Sigma_i$, e.g. from TRPO~\citep{schulman2015trpo}, the resultant action is the belief-weighted sum of mean actions:
\begin{align*}
a^* &= \argmax_a \sum_{i=1}^k b(\phi_i) \log \pi_i(a | s) \\
&= \left[ \sum_{i=1}^k b(\phi_i) \Sigma_i^{-1} \right]^{-1} \sum_{i=1}^k b(\phi_i) \Sigma_i^{-1} \mu_i
\end{align*}

\begin{figure}
  \centering
  \begin{subfigure}{1\linewidth}
    \centering
    \includegraphics[width=1\linewidth]{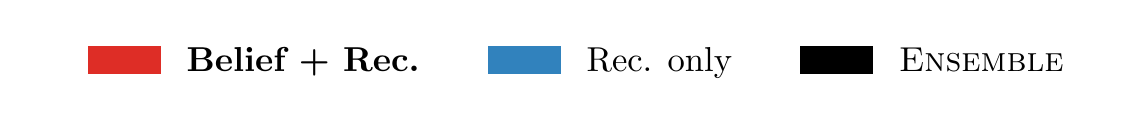}
  \end{subfigure}
  \\
  \begin{subfigure}{0.30\linewidth}
    \centering
    \includegraphics[width=1\linewidth]{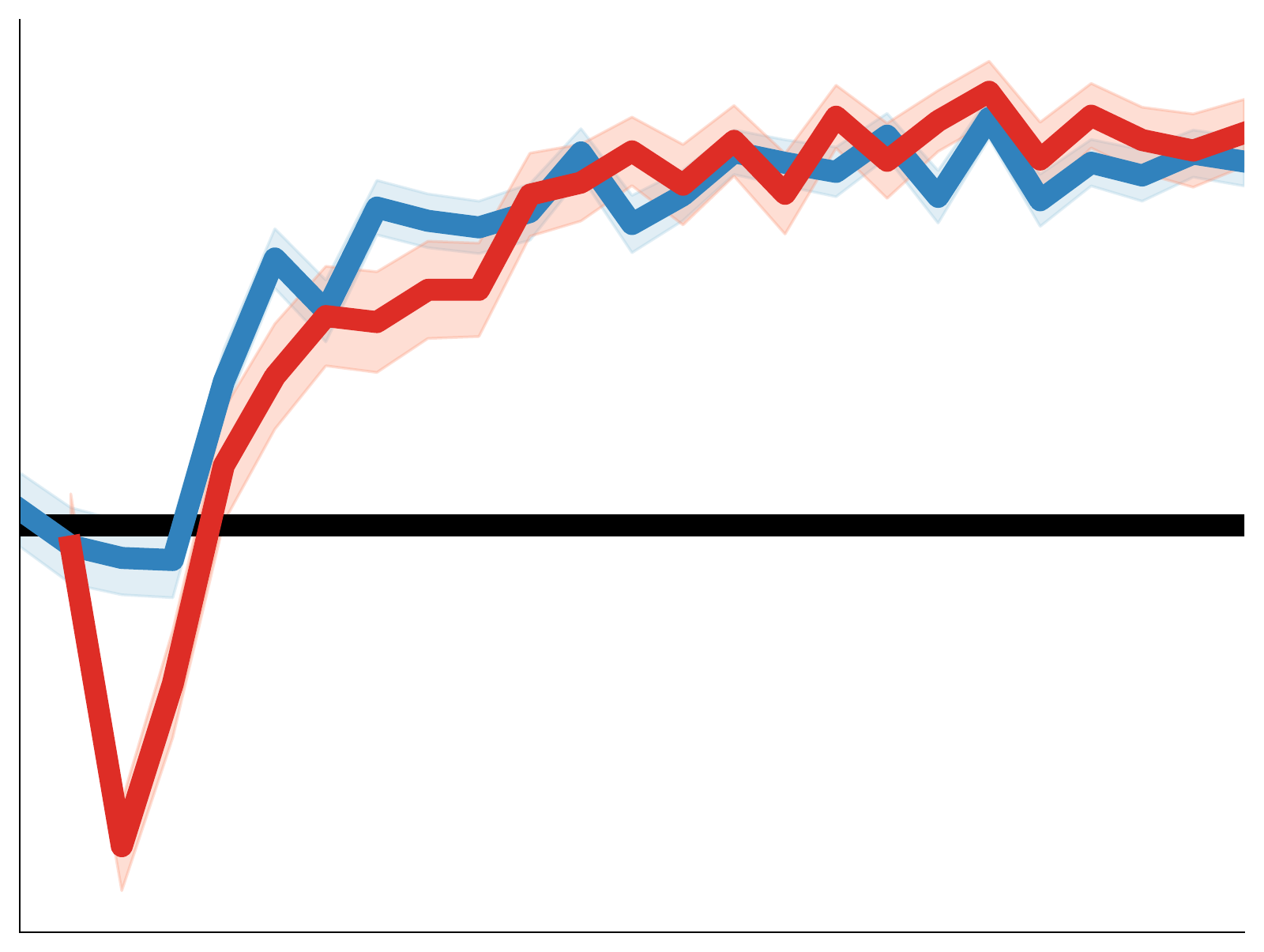}
    \caption{\envMaze}
  \end{subfigure}
~
  \begin{subfigure}{0.30\linewidth}
    \centering
    \includegraphics[width=1\linewidth]{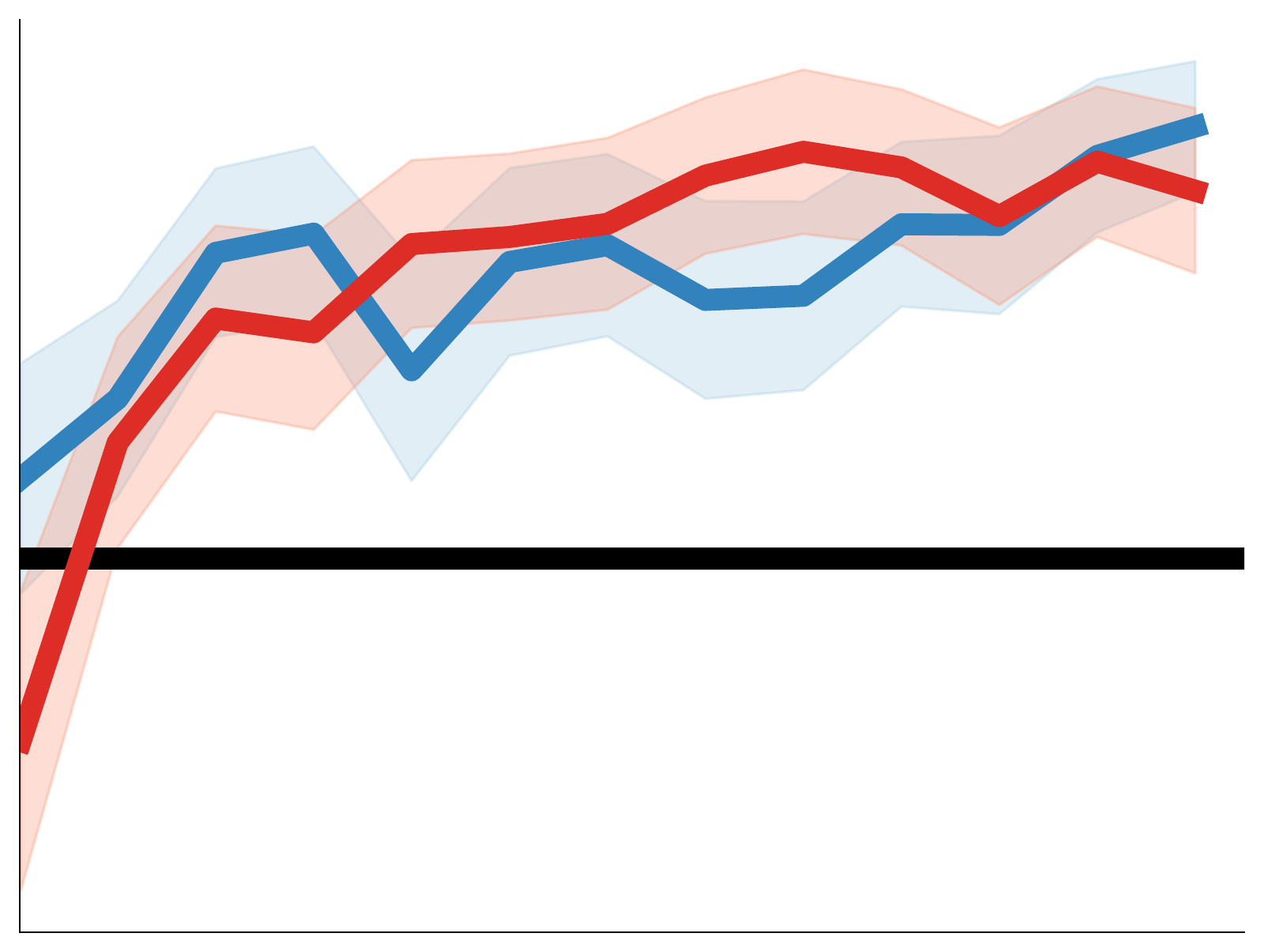}
    \caption{\envMazeTen}
  \end{subfigure}
~
  \begin{subfigure}{0.30\linewidth}
    \centering
    \includegraphics[width=1\linewidth]{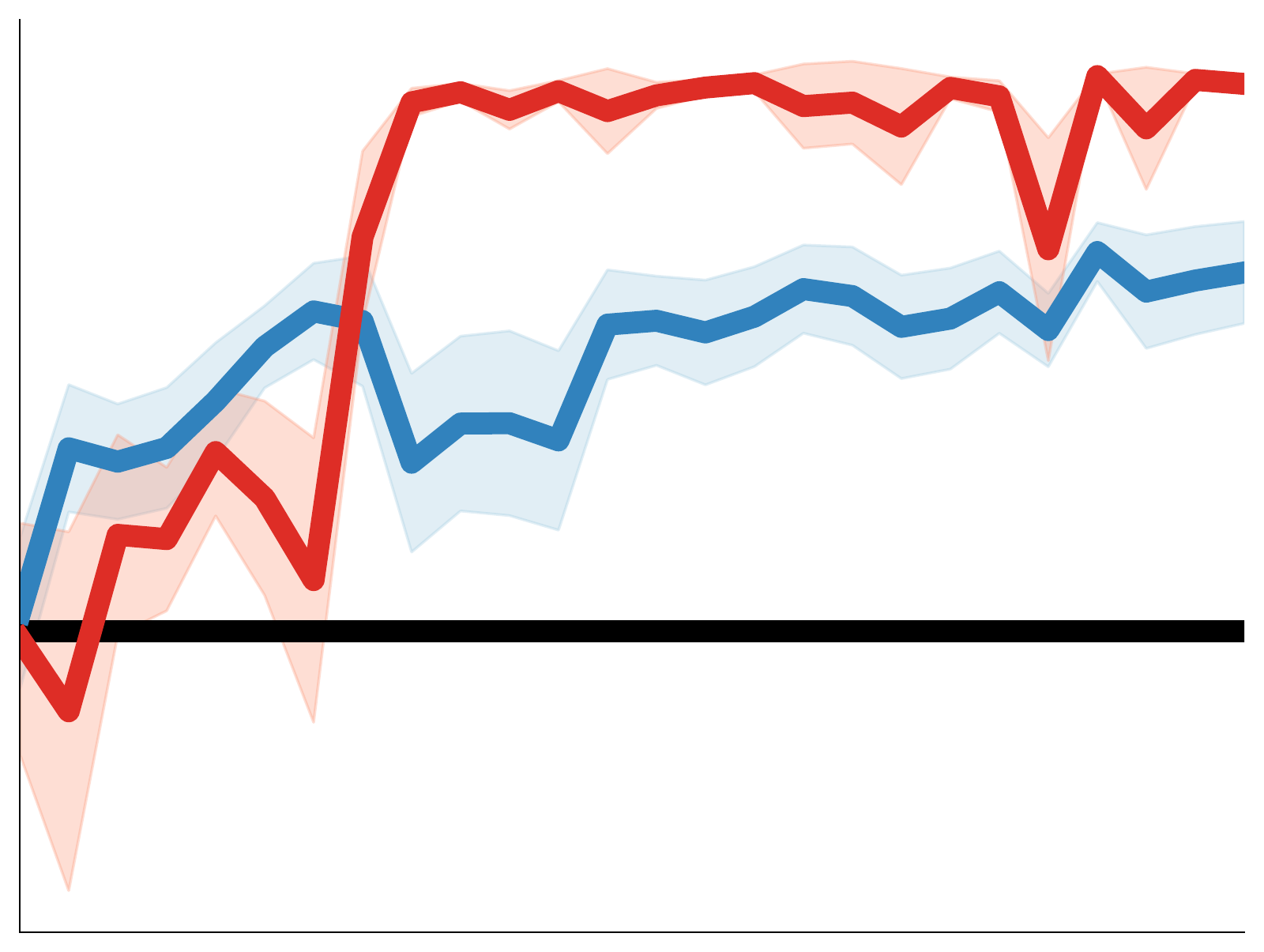}
    \caption{\envDoor}
  \end{subfigure}
  \caption{
    Ablation study on input features. Including both belief and recommendation as policy inputs results in faster learning in \envDoor.
  }
  \label{fig:input-features}
  \end{figure}
\subsection{Ablation Study: Residual Policy Inputs}
\label{app:ablation-input}

The \algName policy takes the belief distribution, state, and ensemble recommendation as inputs (\figref{fig:network}).
We considered two versions of BRPO with different inputs - only recommendation (which implicitly encodes belief), and one with both recommendation and belief.


The results show that providing both belief and recommendation as inputs to the policy are important (\figref{fig:input-features}).
Although \algName with only the recommendation performs comparably to \algName with both inputs on \envMaze and \envMazeTen, the one with both inputs produce faster learning on \envDoor.

\subsection{Ablation Study: Better Sensing Ensemble}
\label{app:ablation-ensemble}

The ensemble we used for training \algName in \figref{fig:benchmark} randomly senses with probability 0.5.
A more effective sensing ensemble baseline policy could be designed manually, and used as the initial policy for the \algName agent to improve on.
Note that in general, designing such a policy can be challenging:
it requires either task-specific knowledge, or solving an approximate Bayesian RL problem.
We bypass these requirements by using \algName.

On the \envMazeTen environment, we have found via offline tuning that a more effective ensemble baseline agent senses only for the first 150 of 750 timesteps.
\tabref{tab:better-ensemble} shows that \algName results in higher average return and success rate.
The performance gap comes from the suboptimality of the ensemble recommendation, as experts are unaware of the penalty for reaching incorrect goals.

\begin{table}[h!]
  \centering
\footnotesize
  \begin{tabulary}{1\linewidth}{L*{5}{R@{$~\pm~$}R}} \toprule
    & \multicolumn{2}{c}{\bf \algName}
    & \multicolumn{2}{c}{RandomSensing}
    & \multicolumn{2}{c}{BetterSensing} \\ \midrule
  Avg. Return & \bf  465.7 & 4.7              & 409.5 &  10.8         & 416.3 &  9.4 \\
  Success Rate&  \multicolumn{2}{c}{\bf 100\%} & \multicolumn{2}{c}{-}&  \multicolumn{2}{c}{96.3\%}\\
  \bottomrule
  \end{tabulary}
  \vspace{1em}
  \caption{
    Comparison of \algName and ensembles on \envMazeTen.}
  \label{tab:better-ensemble}
\end{table}

\end{document}